\documentclass[10pt,twocolumn]{article}

\pdfoutput=1

\usepackage{geometry} 
\usepackage{setspace}
\usepackage{booktabs}

\usepackage{cite}

\usepackage{authblk}
\usepackage{xcolor}

\usepackage{stix2}

\usepackage{amsmath}
\usepackage{bm}
\usepackage[makeroom]{cancel}

\usepackage{siunitx}

\usepackage[english]{babel}

\usepackage{graphicx}
\graphicspath{{figures/}}
\usepackage[font=small]{caption}
\usepackage{subfig}
\usepackage{stfloats} 

\makeatletter
\newcommand*{\centerfloat}{%
	\parindent \z@
	\leftskip \z@ \@plus 1fil \@minus \textwidth
	\rightskip\leftskip
	\parfillskip \z@skip}
\makeatother

\usepackage[colorlinks=true,
linkcolor=blue,
citecolor=blue,
urlcolor=blue,
filecolor=blue,
pdfstartview=FitH]{hyperref}

\definecolor{darkred}{rgb}{0.8500, 0.3250, 0.0980}  
\definecolor{mycyan}{rgb}{0.3010, 0.7450, 0.9330} 
\definecolor{darkblue}{rgb}{0 0.4470 0.7410}
\definecolor{darkyellow}{rgb}{0.9290 0.6940 0.1250} 	
\definecolor{gray}{rgb}{0.60 0.60 0.60} 	
\definecolor{matlabyyLeft}{rgb}{0 0.4471  0.7412}
\definecolor{matlabyyRight}{rgb}{0.8510 0.3255 0.0980}

\definecolor{lightcyan}{RGB}{200,255,255}
\definecolor{lightgreen}{RGB}{225,255,225}
\definecolor{lightblue}{RGB}{225,225,255}
\definecolor{magenta}{RGB}{225,0,225}

\newcommand{\diff}{\,\mathrm{d}}
\newcommand{\pd}[2]{\ensuremath{ \frac{\partial #1}{\partial #2} } }

\newcommand{\unitvector}[1]{\hat{\mathbf{#1}}}
\newcommand{\uvect}[1]{\hat{\mathbf{#1}}}
\newcommand{\vect}[1]{\mathbf{#1}}

\usepackage{oplotsymbl}

\newcommand{\beginsupplement}{%
	\setcounter{section}{0}
	\renewcommand{\thesection}{S\arabic{section}}  
	\renewcommand{\thesubsection}{S\arabic{section}.\arabic{subsection}} 	
	\setcounter{figure}{0}
	\renewcommand{\thefigure}{S\arabic{figure}}%
	\setcounter{table}{0}
	\renewcommand{\thetable}{S\arabic{table}}%
	\clearpage
	\setcounter{page}{1}
}

\newcommand{\etal}{\textrm{et al.~}}

\begin{document}
	
	\title{\textbf{Design optimization and robustness analysis of rigid--link flapping mechanisms}}
	\date{}
	\author[1]{Shyam Sunder Nishad}		
		\author[1]{Anupam Saxena}
	
	\affil[1]{\normalsize Department of Mechanical Engineering, IIT Kanpur, 
		India 208016.}
	\affil[$\dagger$ ]{\small Corresponding author: \url{shyam@iitk.ac.in} }

		\maketitle

	\begin{abstract}	
		Rigid link flapping mechanisms remain the most practical choice for flapping wing micro-aerial vehicles (MAVs) to carry useful payloads and onboard batteries for free flight due to their long-term durability and reliability. However, MAVs with these mechanisms require significant weight reduction to achieve high agility and maneuverability. One approach involves using single-DOF planar rigid linkages, which are rarely optimized dimensionally for high lift and low power, considering their sweeping kinematics and the unsteady aerodynamic effects. We integrated a mechanism simulator based on a quasistatic nonlinear finite element method with an unsteady vortex lattice method-based aerodynamic analysis tool within an optimization routine. We optimized three different mechanism topologies from the literature. Significant power savings were observed up to 34\% in some cases, due to increased amplitude and higher lift coefficients resulting from optimized asymmetric sweeping velocity profiles. We also conducted a robustness analysis to quantify performance sensitivity to manufacturing tolerances. It provided a trade-off between performance and reliability and revealed the need for tight manufacturing tolerances and careful material selection. Finally, the analysis helped select the best mechanism topology, as we observed significant variation in sensitivity to manufacturing tolerances and peak input torque values across different topologies for a given design lift value. The presented unified computational tool can find application in flapping mechanism topology optimization, as it can simulate any generic single-DOF planar rigid linkage without supplying kinematics manually.
	\end{abstract}
	

	\section{Introduction}
\label{sec:introduction}

\begin{figure*}[h!]
	\centerfloat
	\includegraphics[scale=0.9]{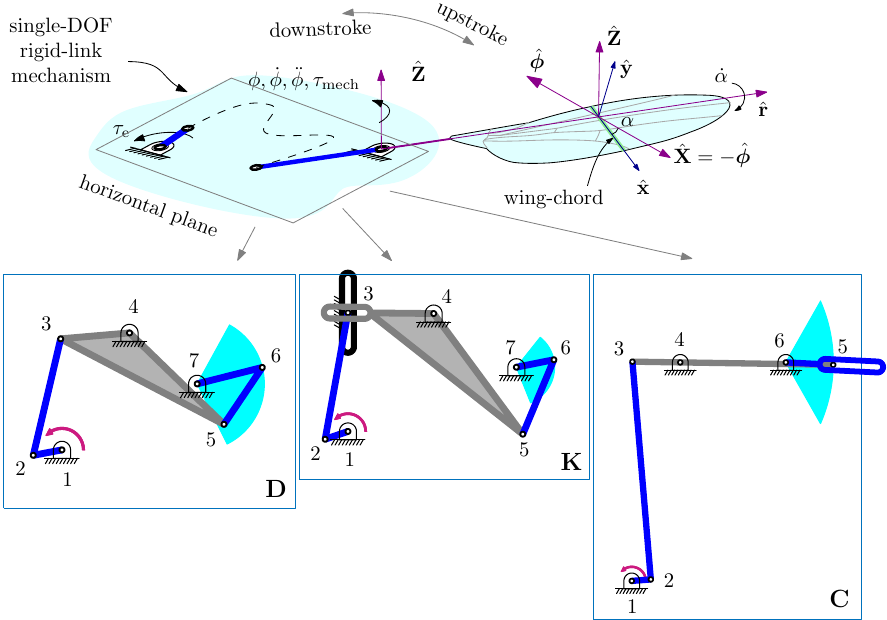}
	\caption{ Optimality of single-DOF planar rigid link flapping mechanisms: Such mechanisms generate the sweeping component of the flapping motion of a thin, flat, rigid wing on a horizontal plane with normal $\hat{\mathbf{Z}}$. The input torque $\tau_\text{e} \hat{\mathbf{Z}} $ results in the torque $\tau_\text{mech} \hat{\mathbf{Z}}$ on the output link connected to the wing whose sweeping angle, velocity, and acceleration are denoted as $\phi$, $\dot{\phi} \hat{\mathbf{Z}} $, $\ddot{\phi} \hat{\mathbf{Z}}$ respectively, and the pitching angle and velocity as $\alpha$ and $\dot{\alpha} \hat{\mathbf{r}}$ respectively. Cylinderical axes $\hat{\bm{\phi}}$ and $\hat{\mathbf{Z}}$ are on the wing-chord plane $\hat{\mathbf{x}}$--$\hat{\mathbf{y}}$. We selected three such mechanisms from the literature for our study, which provide large amplitudes without compromising on the force transmission quality:  Deng \etal \cite{deng2021design} - denoted as D, Karásek \etal \cite{karásek2014pitch} - denoted as K,  Coleman \etal \cite{coleman2017development} denoted as C. The configurations shown here correspond to the minimum force transmissivity index \cite{lin2002force}. The boxes shown here are discussed in Secs. \ref{sec:optimization} and \ref{sec:optimConstraints}. Indices indicated are node numbers used in Secs. \ref{sec:optimization} and \ref{sec:mech_simulator}. The mechanisms were originally optimized for a predefined amplitude and a sinusoidal sweeping angle profile. Some research queries to be explored are: How does one know which amplitude value is optimal? Is restricting the optimization search space to symmetrical sweeping angle profiles justified? }
	\label{fig:question}
\end{figure*}

In the case of flapping wing micro-aerial vehicles (MAVs) with rigid linkage-based mechanisms, the weight becomes significant due to components such as motors, batteries, and flapping mechanisms. However, these mechanisms are still the most practical among all the kinds available \cite{song2023review,singh2022classification,yousaf2021recent}, due to their better long-term durability and reliability when there is a need to carry some useful payload such as a camera along with an onboard battery. For flapping-wing MAVs to achieve high agility, maneuverability, and hovering ability like insects, their weight should be minimized. 
With this idea, most insect-inspired flapping wing MAVs have focused only on the sweeping component of flapping motion, which can be implemented with planar linkages (Fig. \ref{fig:question}). They rely on 
wing flexibility for passive wing pitching, inspired by the same observed in insects \cite{bergou2007passive,whitney2010aeromechanics}. Incorporating active pitching motion in flapping mechanisms in the past has resulted in heavier and more complex designs suitable primarily for benchtop experiments. Stroke plane deviation is often omitted because it negatively 
affects lift and power consumption in most flapping patterns
\cite{hu2019effects,luo2018effects,sane2001control, wang2016numerical}. For further improvements, the single-DOF flapping mechanisms must be optimized to generate high lift and consume low power, thus allowing smaller motors and batteries while maintaining sufficient flight time.

Studies through computational fluid dynamics \cite{zhang2016optimization} and quasi-steady \cite{phan2019extremely}  aerodynamic models have shown that a larger sweeping amplitude reduces the frequency required to achieve a given lift, thus, reducing the inertial power consumption significantly \cite{phan2019extremely}, and consequently the overall power requirement. Larger wings can further help the rigid linkage-based flapping wing MAVs  meet their high lift demands by lowering the power consumption \cite{phan2019extremely}. However, such wings cannot sustain high flapping frequency and work better with larger amplitudes. In line with these observations, six-bar and seven-bar single-DOF rigid-link flapping mechanisms were proposed  \cite{deng2021design,karásek2014pitch, 
	coleman2017development,jeon2017design} that offer large sweeping amplitudes without suffering from poor force transmission quality, unlike the four-bar-based flapping mechanisms (Fig. \ref{fig:question}).
These mechanisms were optimized for a predefined large sweeping amplitude and other kinematic parameters such as symmetry between sweeping angles (and velocity profiles) of the left and right wings or symmetry of upstroke and downstroke velocity profiles, subject to constraints of no quick-return, and transmission angle of each four-bar linkage, Grashof/non-Grashof conditions. However, the effect of the sweeping velocity profile was rarely included in these optimization routines, except for Deng \etal \cite{deng2021design} who aimed to replicate a predefined symmetric sweeping angle profile.  

Wing kinematics optimization studies \cite{berman2007energy,nabawy2015aero,yan2015effects,ke2017wing,nguyen2019neural,bhat2020effects,lang2023sensitivity}, using quasi-steady and unsteady aerodynamic models and experiments, have demonstrated that the mean lift and power consumption during hovering flight vary significantly depending on the sweeping velocity profiles used. This raises an important question: Should the flapping mechanisms be optimized to reproduce a sweeping velocity profile from existing studies? Two arguments challenge this idea. First, while most results agree on the mean lift trends when varying the sweeping velocity profile, there are notable disagreements on the mean power consumption trends due to differences in the aerodynamic models and the exclusion of inertial effects in some studies \cite{nabawy2015aero, lang2023sensitivity, bhat2020effects, yan2015effects}. Second, the sweeping velocity profiles \cite{berman2007energy} used in these studies are symmetrical about (a) the mean stroke position and (b) the two strokes. Restricting a flapping mechanism to such a symmetrical velocity profile could significantly limit the optimization search space. Notably, Terze \etal \cite{terze2021optimized} attempted to address the second problem in their wing kinematics optimization study. However, they excluded aerodynamic power from their analysis, assuming that inertial power consumption was significantly higher, contradicting findings from previous studies \cite{sun2002lift,phan2019extremely}. These observations suggest that instead of optimizing the mechanisms to replicate predefined kinematic parameters, optimization routines need to compute the lift and power consumption in situ by directly simulating the mechanism. This approach would enable the exploration of arbitrary sweeping velocity profiles and flapping amplitudes, unlocking greater design flexibility and potential improvements in performance. However, there are very few studies following this approach.

Huang \etal \cite{huang2019optimization} optimized a double crank-rocker flapping mechanism for a bird-based MAV for lift and thrust. However, they used a quasi-steady aerodynamic model that does not consider the major unsteady effects in flapping flight - the added-mass effect, leading edge vortex effect, and wing-wake interaction. It is based on empirical coefficients, thus having limited prediction ability for optimization purposes. Han \etal \cite{han2023twist} optimized a single-DOF three-dimensional (3D) mechanism for sweeping and pitching motions, to generate high thrust, using the extended UVLM  \cite{nguyen2016extended} that captures the unsteady phenomenon of vorticity shedding from the trailing edge well, along with the added mass effect, wing rotation effect, and wing wake interaction. Yang \etal \cite{yang2024parameter} optimized a planar rigid link folding flapping mechanism using the extended UVLM, considering a weighted lift objective and aerodynamic power coefficients. The effect of amplitude and inertial power was neglected. A  small angle of attack of 6 degrees was adopted, which is insufficient to form a leading-edge vortex - a major contributor to the unsteady aerodynamic force. Further, these optimizations require manually supplying mechanism kinematics and transmission angle information. Therefore, they may not be suitable for applications such as topology selection or optimization of such mechanisms, where generic mechanisms need to be simulated. There is also a need for robustness analysis of the flapping mechanism to ascertain that the mechanism works in reality as per the expected design. Slight variations in link lengths or angles due to manufacturing tolerances, assembly misalignments, or aging may significantly alter the flapping amplitude and kinematics. As aerial vehicles on micro-scales operate at relatively high flapping frequencies, these slight variations can significantly change the aerodynamics, resulting in reduced lift or higher energy consumption.

The present work combines the computation of mechanism kinematics and aerodynamic forces into a single optimization scheme. 
The velocity profile is obtained from a quasistatic nonlinear finite element simulation (FEM) based on the method of truss elements by Rai \etal  \cite{rai2010unified} for a generic single-DOF planar linkage with revolute joints and horizontal/vertical sliding joints extended to (i) include sliders on oscillating ternary links, and (ii) compute the mechanical advantage. Mean lift and power consumption are computed using the blade element theory and a two-dimensional adaptation of the extended UVLM \cite{nguyen2016extended}. The leading-edge vortex effect is modeled using the Polhamus leading-edge suction analogy. A multi-objective optimization is performed that maximizes mean lift and minimizes mean power consumption. The resulting Pareto set combined with a robustness analysis against manufacturing tolerances is used to select the best flapping mechanism (Fig. \ref{fig:pareto}, Tab. \ref{tab:reasonsComparison}).   We demonstrate these by optimizing three different topologies from the literature \cite{deng2021design,karásek2014pitch, coleman2017development}. The presented unified computational tool can also help make informed MAV design decisions based on the payload and flight time requirements from a flapping wing MAV.

The rest of the paper is organized as follows: Sec \ref{sec:optimization} lays out the optimization methodology for a one-DOF planar rigid linkage flapping mechanism. A nonlinear finite element formulation for the mechanism simulator is given in Sec. \ref{sec:mech_simulator}, and details on the aerodynamic analysis are in Secs. \ref{sec:uvlm} -- \ref{sec:power}. Section \ref{sec:results} presents the results, which are discussed in Sec. \ref{sec:discussion} and concluded in Sec. \ref{sec:conclusion}.   	
	\section{Methods}
\label{sec:methods}
\subsection{Mechanism optimization}
\label{sec:optimization}

To simulate a flapping mechanism, we assume the node 1 (Fig. \ref{fig:question}) at the origin. Further, all link lengths are normalized with respect to crank length, such that only the ratio of other lengths is used. Therefore, node 2 is at a unit distance from the node 1. We assume a known initial orientation of the crank so that, for a mechanism with $n$ nodes (revolute joints), we have $2n-4$ coordinates to solve using optimization. 

Using a nonlinear FEM (Sec. \ref{sec:mech_simulator}), we simulate the full rotation of the crank in $M$ equal steps to obtain the rotation angles of the output link in each step as: 
\begin{align*}
	\Delta\phi_1, \Delta\phi_2, \cdots, \Delta\phi_M
\end{align*}
This sequence yields the output link angular velocity $\Omega$ when divided by the time $\Delta t$ for each step. Assuming the crank rotating uniformly with frequency $f$, $\Delta t = 1/(fM)$. The output link acceleration $\dot{\Omega}$ is obtained through finite difference. Amplitude $\phi_0$ can also be obtained from the above sequence.
With this information about the output--link, the aerodynamic force coefficient $C_\text{F}$ is computed using UVLM (Secs. \ref{sec:uvlm} and \ref{sec:force_LEV}), assuming a known wing pitching angle profile $\alpha(t)$. Subsequently, lift and power consumption are computed (Secs. \ref{sec:uvlm} and \ref{sec:power}). 
To compute the transmission quality of the mechanism (Sec. \ref{sec:mech_simulator}), we use the force transmissivity index (FTI) proposed by \cite{lin2002force}. The FTI takes into account the performance of both the input and output links and is better suited for more than four link mechanisms. 

We perform multiobjective optimization of the mechanism to obtain a Pareto set for maximizing lift and minimizing power consumption. For this, the following constraints are enforced: 
\begin{enumerate}
	\item link--length ratio constraint: 
	\begin{align*}
		1 \le \frac{l_{ij}}{l_\text{crank}} \le r_\text{max}
	\end{align*}
	such that, the crank (the input link) is the smallest one. A suitable value of $r_\text{max}$ is selected to prevent very large links from being selected during optimization. 
	The subscripts correspond to node numbers shown in Figs. \ref{fig:question}. 
	\item Grashof criterion to have a crank-rocker linkage on the input side of the flapping mechanism (in absence of crank-slider): 
	\begin{align*} 
		l_{12}  + 2 \max \{l_{23},l_{34},l_{41}\} \le l_{23} + l_{34} + l_{41} + \text{offset}
	\end{align*}
	A small positive value of the offset (0.4: with normalized link lengths) is used to enforce Grashof's criterion strictly, to prevent the mechanism switching from one branch to another. $l_{12}$ is the crank-length.
	\item No quick return condition is imposed as an inequality constraint for optimization, with a small value of $\delta$ (= 0.01), since equality constraints with real numbers are not meaningful. Quick return ratio (QRR):
	\begin{align*}
		1-\delta &< \text{QRR} < 1+\delta, \qquad \text{where,} \\
		\text{QRR} &= \frac{\theta_2 - \theta_1}{2\pi - (\theta_2 - \theta_1) } 
	\end{align*}
	For the crank-rocker part of the flapping mechanisms such as D and C in Fig. \ref{fig:question}, 
	\begin{align*}
		\theta_2 &= \arccos{ \frac{ l_{34}^2 - ( l_{41}^2 + (l_{23} - l_{12})^2  ) }{ 2 l_{41} ( l_{23} - l_{12} ) } } + \pi \\
		\theta_1 &= \arccos{ \frac{ l_{34}^2 - ( l_{41}^2 + (l_{23} + l_{12})^2  ) }{ 2 l_{41} ( l_{23} + l_{12} ) } }
	\end{align*}
	For the crank-slider portion in the mechanism K:
	\begin{align*}
		\theta_2 &= \arccos{ \frac{ x_3 } {  l_{23} - l_{12}  } } + \pi \\
		\theta_1 &= \arccos{ \frac{ x_3 } {  l_{23} + l_{12} } }
	\end{align*}
	If the input part of the mechanism is a crank-slider, we impose $x_3 = x_1 = 0$ for a vertical slider, and $y_3 = y_1 = 0$ for a horizontal slider. Thus, instead of $2n-4$ unknowns, only $2n-5$ need to be solved in this case.
	\item large amplitude 
	\begin{align*}
		\phi_\text{min} \le \phi_0 \le \pi
	\end{align*} 
	
	\item the minimum value of FTI during the entire flapping cycle is the critical value FTI$_\text{cr}$. To avoid singularity and have good transmission quality, it should be high. Therefore, we impose a lower limit on this critical value as:
	\begin{align*}
		\text{FTI}_\text{cr} \ge \text{FTI}_\text{min}
	\end{align*} 
	\item box constraint: the mechanism should lie within a specified polygonal region to restrict its workspace during the entire flapping motion (Fig. \ref{fig:question}). 
\end{enumerate}

\subsection{One-DOF planar linkage simulator}
\label{sec:mech_simulator}

\begin{figure*}[!t]
	\centerfloat
	\includegraphics[scale=0.9]{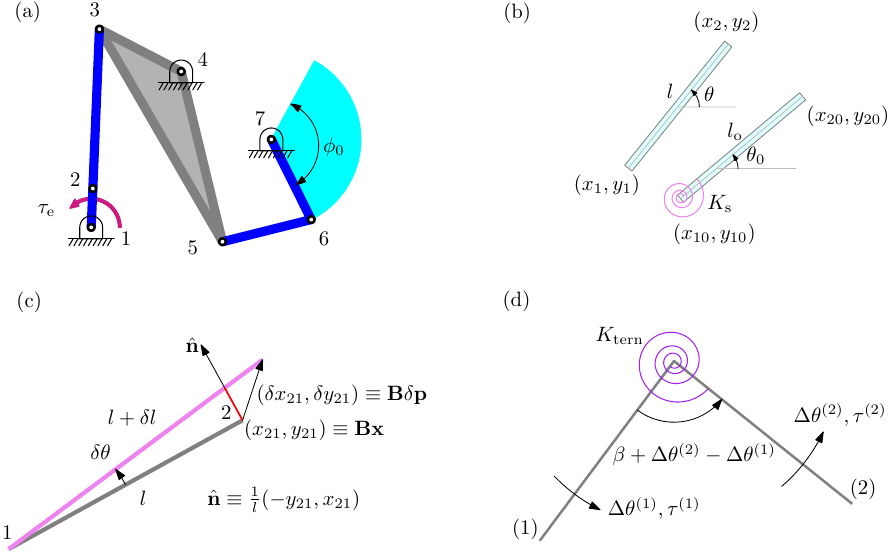}											
	\caption{ Nonlinear FEM based single-DOF planar rigid linkage simulator using truss elements:
		(a)	 The nodes (revolute joints) are numbered 1,2,3, $\cdots$. Joints are considered frictionless. A linear torsional spring with stiffness $K_\text{s}$ (shown in (b)) is attached to one end of each link, with the other end of the spring being grounded.
		(b)	Truss element deformation:   Nodes displace from $(x_{i0},y_{i0})$ to $(x_i, y_i)$, $i=1,2$ under external torque load $\tau_\text{e}$ at the actuated joint, resulting in rotation of the element by $\theta-\theta_0$ and axial deformation $l-l_\text{o}$. 
		(c)	Truss element rotation $\delta\theta$ and stretch $\delta l$ due to the virtual displacement $\delta\mathbf{p}$: $(x_{21}, y_{21})$ and $(\delta x_{21}, \delta y_{21})$ are the current position and the virtual displacement of node 2 relative to node 1, respectively. $\unitvector{n}$ is the normal to the current orientation of the element. Virtual displacements are arbitrarily small while applying the virtual work principle.
		(d)	Ternary link with angle $\beta$: its two truss members (1) and (2) connected through a stiff torsional spring of stiffness $K_\text{tern}$. $\Delta\theta^{(1)}$ and $\Delta\theta^{(2)}$ correspond to the rotation of truss members from their undeformed positions (not shown here).
	}
	\label{fig:linkage_fem}
\end{figure*}

Joints are assumed to be frictionless. To analyze a rigid linkage using the nonlinear finite element method \cite{rai2010unified}, each binary link is modeled as a truss element with large axial stiffness that undergoes little strain but large rotation. A soft linear torsional spring of stiffness $K_\text{s}$ is attached to the input link  (Fig. \ref{fig:linkage_fem}(b)) whose other end is grounded. We attach a similar spring, but of high stiffness, to the output link to compute the mechanical advantage (MA, needed for FTI computation).

We simulate the mechanism in two states: (i) unlocked state in which the mechanism configuration changes and (ii) locked state in which MA is computed. In the unlocked state, we activate only the input link spring. To rotate the input link by an angle $\Delta\theta$, we apply an external torque $\tau_\text{e}= K_\text{s} \Delta\theta$ on the input link and let the system attain equilibrium quasi-statically. In the locked state, we also activate the stiff spring at the output link, which results in little rotation. Then, MA is computed as the ratio of torque developed in the output spring to the input torque. We repeat the two-state process in the successive configurations, increasing the input torque by $K_\text{s} \Delta\theta$ each time. This torque increment essentially resets the equilibrium position $\theta_0$ of the input spring in each step ($\theta_0 = \theta_0 + \Delta\theta$: deduced from Eq. \eqref{eq:virtualwork} by substituting $\tau_\text{e} = K_\text{s} \Delta\theta $), and the effective external load in each step remains $K_\text{s} \Delta\theta$. This effective load lets the input link rotate by $\Delta\theta$ in each step.

We implement a slider on an oscillating binary link by setting its axial stiffness to zero (e.g., link 5--6 joining nodes 5, 6 in mechanism C, Fig. \ref{fig:question}). Using the method by Rai \etal \cite{rai2010unified}, a ternary link could be implemented with three truss elements if there are only revolute joints (e.g., link 3--4--5 in mechanism D). However, if there is a slider on an oscillating ternary link (e.g., link 3--4--5 in mechanism K), lengths of the truss members connected to the slider vary freely because of its zero axial stiffness (like the members 3--4, 3--5), thus, adding an extra DOF and rendering the existing method incapable. Hence, we proposed a new method for such cases. 

To derive the element force vectors and element tangent stiffness matrices, we use the virtual work principle.
Given the nodal positions $\mathbf{x}_0 = \begin{bmatrix}
	x_{10} & y_{10} & x_{20} & y_{20}
\end{bmatrix} ^\text{T}$ of a truss element in an equilibrium configuration, we need to find the new equilibrium position $\mathbf{x} = \begin{bmatrix}
	x_1 & y_1 & x_2 & y_2
\end{bmatrix} ^\text{T}$, when external load torque $\tau_\text{e}$ is applied (Fig. \ref{fig:linkage_fem}(b)). 
The virtual work residual of the internal and external forces can be written as: 
\begin{align}
	\mathbf{g} \cdot \delta\mathbf{x} = 
	\underbrace{\int \sigma \delta\epsilon \diff V}_\text{on truss element} +  \underbrace{K_\text{s} (\theta - \theta_0) \delta \theta}_\text{on spring} - \underbrace{\tau_\text{e} \delta \theta}_\text{by external load}  
	\label{eq:virtualwork}
\end{align}
where, $\sigma$ is the stress in the element due to the displacement $\mathbf{x} - \mathbf{x}_0$ during which the element rotates from initial orientation $\theta_0$ to $\theta$. $\delta\epsilon$ is the strain due to the virtual displacement $\delta\mathbf{x}$. $\delta\theta$ is the corresponding rotation of the element.  
Assuming uniform small strain and volume conservation ($V = l_\text{o} A_\text{o} = l A$), we obtain:
\begin{align}
	\int \sigma \delta\epsilon  \diff V = (E\epsilon) \delta\epsilon (l_\text{o} A_\text{o}) 
	= E A_\text{o} \frac{l - l_\text{o}}{l_\text{o} } \delta l = G  \frac{l - l_\text{o}}{l_\text{o} } \delta l 
	\label{eq:stress_integral}
\end{align}
where, $E$ is the Young's elasticity modulus, $l_\text{o}$ and $l$ are the original and deformed lengths, $A_\text{o}$ and A are the corresponding cross-sectional areas, and $EA_\text{o} = G$ (axial stiffness).
From Fig. \ref{fig:linkage_fem}(c): 
\begin{align}
	l^2  = 	\begin{bmatrix}
		x_{21} &
		y_{21}
	\end{bmatrix}
	\begin{bmatrix}
		x_{21} \\
		y_{21}
	\end{bmatrix}	= \mathbf{x}^\text{T} 
	(\mathbf{B}^\text{T} \mathbf{B} )  
	\mathbf{x}
	\implies
	\delta l &= \frac{1}{l  } \mathbf{y}^\text{T}\delta\mathbf{x}
	\label{eq:delta_l}
\end{align}
where, 
\begin{align*}
	\begin{bmatrix}
		x_{21} \\
		y_{21}
	\end{bmatrix}
	&= 
	\begin{bmatrix}
		x_2 - x_1 \\
		y_2 - y_1
	\end{bmatrix}
	=
	\underbrace{
		\begin{bmatrix}
			-1 & 0 & 1 & 0 \\
			0 & -1 & 0 & 1 
		\end{bmatrix} 
	}_\mathbf{B}
	\mathbf{x} 
	, & \mathbf{y} &= (\mathbf{B}^\text{T} \mathbf{B} )   \mathbf{x} 
\end{align*}
From Fig. \ref{fig:linkage_fem}(c), we can compute the virtual rotation $\delta\theta$ as:
\begin{align}
	\delta \theta &= \frac{ \unitvector{n} \cdot \mathbf{B} \delta\mathbf{x} }{l}
	= \frac{1}{l^2} 
	\begin{bmatrix}
		- y_{21} & x_{21}
	\end{bmatrix}  \mathbf{B} \delta\mathbf{x}
	= \frac{1}{l^2} \mathbf{z}^\text{T} \delta\mathbf{x}
	\label{eq:delta_theta}
\end{align}
where, 
\begin{align*}
	\mathbf{z} = \mathbf{B}^\text{T} \begin{bmatrix}
		- y_{21} \\ x_{21} 
	\end{bmatrix}
	= \mathbf{B}^\text{T} 
	\underbrace{ 
		\begin{bmatrix}
			0 & 1 & 0 & -1 \\
			-1 & 0 & 1 & 0 
		\end{bmatrix}
	}_\mathbf{C}
	\mathbf{x}  
\end{align*}
Substituting in Eq. \eqref{eq:virtualwork} from Eqs. \eqref{eq:stress_integral}--\eqref{eq:delta_theta}, and invoking arbitrariness of the virtual displacement $\delta\mathbf{x}$, we obtain:
\begin{align}
	\mathbf{g}(\mathbf{x}) = {a} \mathbf{y}  +
	{b}  \mathbf{z} 
	\label{eq:f_int}
\end{align}
From this equation, the tangent stiffness matrix $\mathbf{K}_\text{t}$ for the element can be obtained as: 
\begin{align}
	\mathbf{K}_\text{t} = \pd{\mathbf{g} }{\mathbf{p}}= c \mathbf{y}  \mathbf{y}^\text{T}      + a  \mathbf{B}^\text{T} \mathbf{B}    + b \mathbf{B}^\text{T} \mathbf{C}  
	-e  \mathbf{z}\mathbf{y}^\text{T}    + d  \mathbf{z} \mathbf{z}^\text{T}
\end{align}
where, 
\begin{align}
	a &= \frac{G}{l} \frac{l - l_\text{o} }{l_\text{o}},  &  b &=  \frac{   K_\text{s} (\theta - \theta_0) - \tau_\text{e}   }{l ^2}, & \nonumber \\ 
	c &= \frac{G}{l^3}, & d &= \frac{K_\text{s}}{l^4}, & e &=  \frac{2 b}{l^2} 
	\label{eq:coeffs_a_b}
\end{align}

A ternary link is implemented with two truss members with a stiff linear torsional spring of stiffness $K_\text{tern}$ between them (Fig. \ref{fig:linkage_fem}(d)). The high stiffness of this spring ensures that the angle $\beta$ between the two members changes very little. As the truss members deflect by angles $\Delta\theta^{(1)}$ and $\Delta\theta^{(2)}$ from their undeformed positions, they experience the torque $\tau^{(1)}$ and $\tau^{(2)}$  respectively due to this spring: 
\begin{align*}
	\tau^{(1)} &= -K_\text{tern} \left(\Delta\theta^{(1)} - \Delta\theta^{(2)} \right), & 
	\tau^{(2)} &= -K_\text{tern} \left(\Delta\theta^{(2)} - \Delta\theta^{(1)} \right) 
\end{align*}
Superscripts (1) and (2) correspond to the truss members 1 and 2 of the ternary link.
Thus, the virtual work equation in Eq. \eqref{eq:virtualwork} for these two members gets modified to:
\begin{align*}
	\mathbf{g}^{(1)} \cdot \delta\mathbf{x}^{(1)} = \int \sigma^{(1)} \delta\epsilon^{(1)} \diff V
	+  \left(K_\text{s} \Delta\theta^{(1)}  - \tau_\text{e}^{(1)}\right) \delta \theta^{(1)}
	\nonumber\\
	+ K_\text{tern} \left(\Delta\theta^{(1)} - \Delta\theta^{(2)} \right)  \delta \theta^{(1)} \\
	\mathbf{g}^{(2)} \cdot \delta\mathbf{x}^{(2)} = \int \sigma^{(2)} \delta\epsilon^{(2)} \diff V
	+  \left(K_\text{s} \Delta\theta^{(2)} - \tau_\text{e}^{(2)}\right) \delta \theta^{(2)}
	\nonumber\\
	+ K_\text{tern} \left(\Delta\theta^{(2)} - \Delta\theta^{(1)}\right) \delta \theta^{(2)}		
\end{align*}
The spring $K_\text{s}$ and external torque $\tau_\text{e}$ are retained for generality. Solutions are similar to  Eq. \eqref{eq:f_int}, but depend on the nodal positions of both members, due to coupling through the stiff spring.
\begin{align*}
	\mathbf{g}^{(1)}( \mathbf{x}^{(1)}, \mathbf{x}^{(2)} ) = a^{(1)} \mathbf{y}^{(1)} + b^{(1)} \mathbf{z}^{(1)} + {b'}^{(1)} \mathbf{z}^{(1)}  \\
	\mathbf{g}^{(2)}( \mathbf{x}^{(1)}, \mathbf{x}^{(2)} ) = a^{(2)} \mathbf{y}^{(2)} + b^{(2)} \mathbf{z}^{(2)} + {b'}^{(2)} \mathbf{z}^{(2)} 
\end{align*}
Thus, we obtain tangent stiff matrix terms for the truss member 1 from both $\pd{\mathbf{g}^{(1)}}{\mathbf{x}^{(1)}}$ and $\pd{\mathbf{g}^{(1)}}{\mathbf{x}^{(2)}}$, where,
\begin{align}
	\pd{\mathbf{g}^{(1)}}{\mathbf{x}^{(1)}} &= c^{(1)} \mathbf{y}^{(1)} {\mathbf{y}^{(1)}}^\text{T} + a^{(1)} \mathbf{B}^\text{T} \mathbf{B} + \left(b^{(1)}+{b'}^{(1)}\right) \mathbf{B}^\text{T} \mathbf{C} 
	\nonumber
	\\ - &\left(e^{(1)} + {e'}^{(1)} \right) \mathbf{z}^{(1)} {\mathbf{y}^{(1)}}^\text{T} + \left(d^{(1)} + {d'}^{(1)}\right) \mathbf{z}^{(1)} {\mathbf{z}^{(1)}}^\text{T}
	\label{eq:tern_K11}
	\\
	\pd{\mathbf{g}^{(1)}}{\mathbf{x}^{(2)}} &= - \frac{K_\text{tern}}{{l^{(1)}}^2 {l^{(2)}}^2} \mathbf{z}^{(1)} {\mathbf{z}^{(2)}}^\text{T}
	\label{eq:tern_K12}
\end{align}
The expressions for $a^{(1)}$, $a^{(2)}$, $b^{(1)}$, $b^{(2)}$, $c^{(1)}$, $c^{(2)}$, $d^{(1)}$, $d^{(2)}$ are obtained from  Eq. \eqref{eq:coeffs_a_b} by substituting values for the respective truss members, whereas, 
\begin{align*}
	{b'}^{(1)} &= \frac{K_\text{tern}}{{l^{(1)}}^2} \left(\Delta\theta^{(1)} -\Delta\theta^{(2)} \right),  & 
	{b'}^{(2)} &= \frac{K_\text{tern}}{{l^{(2)}}^2} \left(\Delta\theta^{(2)} -\Delta\theta^{(1)} \right) 
	\nonumber\\
	{d'}^{(1)} &= \frac{K_\text{tern}}{{l^{(1)}}^4}, & {e'}^{(1)} &= \frac{2 {b'}^{(1)}}{{l^{(1)}}^2} 
\end{align*}
Tangent stiffness matrix terms for the truss member 2 can be obtained by interchanging indices 1 and 2 in Eqs. \eqref{eq:tern_K11} and \eqref{eq:tern_K12}.

In indicial notation, the terms of global tangent stiffness matrix can be written as:
\begin{align*}
	K_{ij} &= \pd{g_i}{x_j}, &
	\text{since, }
	\delta g_i &= \pd{g_i}{x_j}  \delta x_j = K_{ij} \delta x_j
\end{align*}
When assembling element tangent stiffness matrices $\pd{\mathbf{g}}{\mathbf{x}}$ to form the global tangent stiffness matrix $K_{ij}$, the terms from $\pd{\mathbf{g}}{\mathbf{x}}$ go in the rows corresponding to the global DOFs of the element whose force residual vector $\mathbf{g}$ is used, and in columns corresponding to the global DOFs of the element whose position vector $\mathbf{x}$ is used.  Thus, in the case of ternary links, 	 
$\pd{\mathbf{g}^{(1)}}{\mathbf{x}^{(1)}}$ corresponds to global DOFs from the truss member 1 alone. But, in case of $\pd{\mathbf{g}^{(1)}}{\mathbf{x}^{(2)}}$, the rows correspond to those from truss member 1, while columns to the truss member 2. Similarly,  the element stiffness matrices $\pd{\mathbf{g}^{(2)}}{\mathbf{x}^{(2)}}$ and $\pd{\mathbf{g}^{(2)}}{\mathbf{x}^{(1)}}$ for the truss member-2 of the ternary link, are assembled. 
Once the global stiffness matrix and global force residual vector are formed, Newton Raphson method is used to solve for the nodal positions $\mathbf{x}$, after removing their rows and columns corresponding to the fixed DOFs, as usual.

We compute the force transmissivity index (FTI) as \cite{lin2002force}: 
\begin{align*}
	\text{FTI} = | \text{MA} \times \text{EFR} |
\end{align*}
where, MA is the mechanical advantage of the linkage. EFR, the effective force ratio, is the ratio of power flowing in the output link from the input side of the mechanism to the maximum power that can flow from the same side. 		
In case there is only one link connected to the output link from the input side, as in Fig. \ref{fig:linkage_fem} and the output link is connected to the ground, 
\begin{align*}
	\text{EFR} = \sin\gamma = \frac{  g_2 x_{21} - g_1 y_{21}  }{  \sqrt{g_1^2 + g_2^2} \sqrt{x_{21}^2 + y_{21}^2}  } 
\end{align*}
where, $\gamma$ is the angle between the output link and the force $(g_1,g_2)$ applied by the connected link on the output link. Here, the vector $(x_{21}, y_{21})$ gives the orientation of the output link (Fig. \ref{fig:linkage_fem}C). The internal force  $(g_1,g_2)$ is obtained from the  $4\times 1$ element force vector $\mathbf{g}$ corresponding to the connected link with the output spring active. 
The mechanical advantage is obtained as:
\begin{align*}
	\text{MA} = \frac{K_\text{out} \Delta\phi}{ \tau_\text{e} }
\end{align*}
where, $\Delta\phi$ is the output link rotation computed during the locked--state in each configuration. $K_\text{out}$ is the  stiffness of the output spring.

\subsection{Aerodynamic model}
\label{sec:uvlm}

For the aerodynamic analysis, we use blade element theory to divide the wing into thin chords and analyze a wing chord using the 2D unsteady vortex lattice method (UVLM). UVLM is based on the potential flow theory and provides a good balance between computational cost and accuracy. It cannot inherently model the leading edge separation (LEV) and viscous effects. However, it captures the unsteady phenomenon of vorticity shedding from the trailing edge well, along with the added mass effect, wing rotation effect, and wing wake interaction. 
Dickinson \etal \cite{dickinson1999wing} showed experimentally that the aerodynamic forces in hovering flapping wing motion are roughly normal to the wing, indicating that viscosity affects majorly in structuring vorticity around the wing rather than the aerodynamic forces. Persson \etal \cite{persson2012numerical}  demonstrated through numerical simulations that the potential theory-based numerical solvers can give the correct trend of forces when flow separated at leading edge reattaches and is suitable for preliminary design purposes. Ansari \etal \cite{ansari2006non1, ansari2006non2} modeled vorticity shedding from both leading and trailing edges in their 2D discrete vortex model based on potential theory. However, their approach is computationally costlier than UVLM, as it involves solving two nonlinear integral equations compared to a system of linear equations in UVLM, thus prohibiting its use in optimization problems due to a large number of function evaluations. Roccia \etal \cite{roccia2013modified} included vorticity shedding from both leading and trailing edges in UVLM. However, that presented numerical instabilities due to wing-wake interaction in hovering. Therefore,
Nguyen \etal \cite{nguyen2016extended} incorporated the LEV effect through Polhamus leading-edge suction analogy along with a vortex core growth model for wake vortices to address this problem and validated UVLM results against CFD and experimental data. 
Wang \etal \cite{wang2004unsteady} reported that the unsteady force estimates from 2D CFD computations are not much different from experiments because insects reverse their strokes before a leading edge vortex would shed in a corresponding 2D wing motion. In light of the above evidences, we chose the extended UVLM formulation by Nguyen \etal \cite{nguyen2016extended} for our aerodynamic analysis but traded off the improved accuracy of 3D computations for higher computation speed with its 2D-variant.

\begin{figure*}[h!]
	\centerfloat
	\includegraphics[scale=0.9]{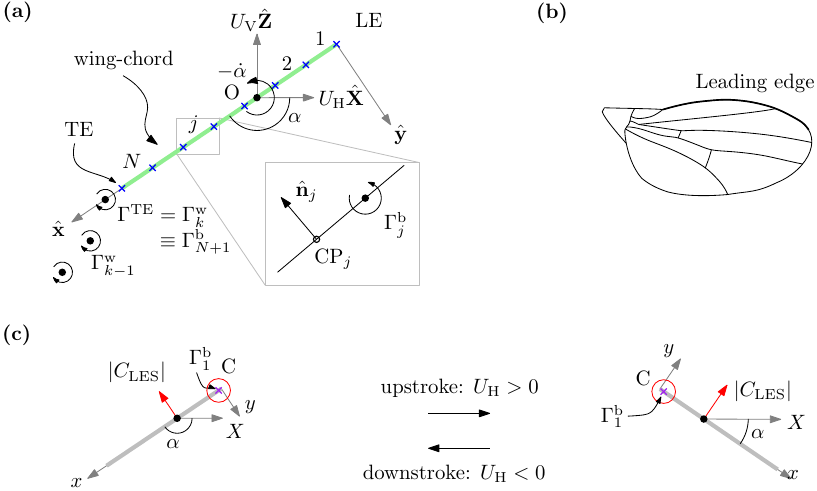}				
	\caption{ Aerodynamic model: 
		\textbf{(a)} 2D-UVLM:  
		A thin, flat, rigid wing--chord with orientation $\alpha$ relative to the horizontal (Fig. \ref{fig:question}), translates with velocity $U_\text{H} \uvect{X}$, and deviates from the horizontal stroke plane with velocity $U_\text{V} \uvect{Z}$. It also rotates with velocity $ \dot{\alpha}\unitvector{r}$ about O. Due to the wing motion, a bound circulation develops around the wing chord. In 2D UVLM, the wing-chord is split into $N$ panels (elements) of length $d$ each, numbered $1,2, \cdots, j, \cdots, N$. Using the lumped vorticity model \cite{katz2001low}, circulation 
		$\Gamma^\textrm{b}_j$ in each panel is placed at $d/4$ from the end towards the leading edge LE. No-penetration condition is imposed at $3d/4$ (collocation points CP$_j$)  from the same end. A trailing edge vortex (TEV) with circulation $\Gamma^\text{TE}(\equiv \Gamma^\text{b}_{N+1}$: notation) is assumed bound at a distance $d/4$ from the trailing edge TE, which sheds into the wake at each time instant. $\Gamma^\text{w}_k$ is the TEV shed at the $k^\text{th}$ time step $t_k$ ($k=1$: the initial step). That means, for analysis at $t_k$, there are $N_\text{w}=k-1$ vortices in the wake, and $\Gamma^\text{w}_k = \Gamma^\text{TE}$  is still bound. The TEV and all wake vortices are transported with the flow, retaining their circulations constant at all times, as per Kirchoff's equation of motion of free vortices. Total circulation ($\sum_{j=1}^{N+1} \Gamma^\text{b}_{j} + \sum_{i=1}^{N_\text{w}}\Gamma^\text{w}_i$) is conserved at each time instant as per Kelvin's theorem.  
		\textbf{(b)} 
		Fruit fly wing for force computations: A simplified design is considered for inertial force computation: The skeleton is assumed to be made of solid carbon fiber rods of 0.8 mm diameter. Six straight rods of length equal to the wing length and two small rods of length ${\frac{1}{5}}^\text{th}$ of the mean wing chord are assumed to make the skeleton. To simplify computations, the mass in these rods is uniformly distributed over the entire wing-surface area. The wing membrane is assumed to be made of a mylar sheet of 10 microns thick. 
		\textbf{(c)} Leading edge vortex (LEV) force using the Polhamus leading-edge suction analogy: Since the LEV exists only on the upper side of the wing chord throughout the translation phase, the force is always directed upwards. 
		C is the contour loop centered at the leading edge panel vortex $\Gamma^\text{b}_1$, used to compute the LES force using the Blasius theorem. }  
	\label{fig:aerodynamic}
\end{figure*}

A brief summary of the 2D UVLM is given in Fig. \ref{fig:aerodynamic}a, and more details can be found in App. \ref{app:uvlm} and Ref. \cite{katz2001low}. 
The nondimensional 2D-UVLM yields the force coefficient $C_\text{F}$ for 2D motion of wing-chord given by nondimensional sweeping velocity $\tilde{\Omega}(\tilde{t})$ and pitching angle $\alpha(\tilde{t})$, where $\tilde{t}$ is the nondimensional time. In inviscid flow, the force coefficient $C_\text{F}$ remains the same for all the wing chords of the finite wing, neglecting the tip effect.
Therefore, using blade element theory, the lift for a finite wing of length $R$ (surface area $S$) and second moment of area $r_2$, flapping at frequency $f$ is obtained as:
\begin{align}
	L 
	= C_\text{L} \frac{\rho}{2} \Omega_\text{ref}^2  r_2^2 S =  C_\text{L} \frac{\rho}{2} \Omega_\text{ref}^2  \hat{r}_2^2 R^2 S 
	\label{eq:lift}
\end{align}
where,  $\hat{r}_2 = r_2/R$ is the nondimensional second moment of area.
The aerodynamic torque in the plane of the motion of the mechanism, assuming horizontal stroke plane, is computed from the horizontal force: 
\begin{align*}
	\bm{\tau}_\text{aero} = \int r \uvect{r} \times C_\text{H} \frac{\rho}{2} \Omega_\text{ref}^2 r^2 \diff S  \uvect{\bm{\phi}}
	= C_\text{H} \,  \frac{\rho}{2}\Omega_\text{ref}^2  r_3^3 S \uvect{Z} 
\end{align*}
where, ${r}_3$ is the third moment of the wing area, lift coefficient $C_\text{L} = C_\text{F}  \cos\alpha$, horizontal force-coefficient $C_\text{H} = -C_\text{F} \sin\alpha$, and $\Omega_\text{ref} = 2\phi_0\, f$, $\phi_0$ being the sweeping amplitude, and $\rho$, the fluid density (App. \ref{app:uvlm}).

\subsection{Force due to LEV}	
\label{sec:force_LEV}

We use the Polhamus leading-edge suction analogy \cite{polhamus1971predictions} to estimate the force due to the leading edge vortex (LEV). According to this analogy, the force due to the leading edge vortex in a separated flow is equal in magnitude to the suction force if the flow remains attached. However, the direction of the force in a separated flow is rotated normally to the wing chord. Therefore, this force coefficient $C_\text{LES}$ is added to the normal force coefficient  $C_\text{F}$, when the angle of attack $\alpha$ is greater than a critical value $\alpha_\text{c}$, above which the leading edge separation occurs (Fig. \ref{fig:aerodynamic}c). The suction force may be less than the theoretically
predicted value due to viscous effects and the effects of leading-edge shape \cite{nguyen2016extended, bartlett1955experimental}. Therefore, a factor $\eta \in (0,1)$ is included.
\begin{align*}
	C_\text{F,net} = C_\text{F} + \eta \, \text{sign}(\cos\alpha) \, | C_\text{LES} |
\end{align*}
When $\alpha <\alpha_\text{c}$, no separation occurs. Hence, $C_\text{LES}$ is parallel to the wing-chord, directed towards the leading edge, and added vectorially to $C_\text{F} \,\unitvector{y}$. In this paper, $\alpha_\text{c} = 12$ deg is taken, similar to that by Nguyen \etal \cite{nguyen2016extended} and Roccia \etal \cite{roccia2013modified}. It is also noted that the expression for suction force considers only the steady terms. During the rotation phase of the flapping motion, the role of LEV becomes less significant compared to the rotational effects. Hence, the error due to the exclusion of unsteady terms may not be significant (similar assumption in \cite{nguyen2016extended}); Roccia \etal \cite{roccia2013modified} do not use a suction model as they consider shedding of vortices from both the leading and trailing edges).

The leading edge suction force is computed using application of the Kutta--Joukowski expression on the leading edge panel vortex (LEPV) with circulation $\Gamma_1^\text{b}$:
\begin{align*}
	F_\text{x}  &= \rho\, q_\text{y}\, \Gamma_1^\text{b}, & F_\text{y} &= - \rho\, q_\text{x}\, \Gamma_1^\text{b}  
\end{align*}
Here, $q_\text{x}$ and $q_\text{y}$ are the flow velocities relative to the wing at the location of LEPV , after excluding the singularity due to LEPV itself.
Note that the normal force $F_\text{y}$ is already included in the force computed in UVLM from  Eq. \eqref{eq:force_ceoff}. $F_\text{x}$ is the leading edge suction force. This can be nondimensionalized to obtain the corresponding force coefficient as:
\begin{align*}
	C_\text{LES}  = \frac{F_\text{x}}{\frac{1}{2} U_\text{ref}^2\, \bar{c}} = 2 \tilde{q}_y \tilde{\Gamma}^\text{b}_1
\end{align*}

\subsection{Computation of inertial torque and power}
\label{sec:power}

We consider a fruit fly wing planform for our aerodynamic computations and use a simplified structure of the wing for inertial power computation, as detailed in Fig. \ref{fig:aerodynamic}b. If the sweeping acceleration is $\dot{\Omega}$, the inertial torque on the wing can be written as: 
\begin{align*}
	\tau_\text{inertial} &= I  \dot{\Omega}, & I &= \rho_\text{w} S \, t_\text{w} r_2^2 
\end{align*}
where, $I$ is the mass moment of inertia of the wing with second moment of area $r_2$, 
$S = {R^2}/{r_\text{a}}$ being the wing-surface area, $R$, $r_\text{a}$, $t_\text{w}$ and $\rho_\text{w}$ -- length, aspect--ratio, thickness and density of the wing respectively.
The wing--density $\rho_\text{w}$ is estimated by adding contribution of mass $m_\text{c}$ of wing--skeleton formed from carbon--fiber rods (distributing uniformly over wing--area) to the wing membrane density $\rho_\text{mylar}$:  
\begin{align*}
	\rho_\text{w} &= \frac{m_\text{c} }{S \,t_\text{w}}  + \rho_\text{mylar}, & m_\text{c} &= l_\text{c} \, \pi r_\text{c}^2 \,  \rho_\text{c}, & l_\text{c} &=  6 R + \frac{2}{5} \frac{R}{r_\text{a}} 
\end{align*}
where,  $l_\text{c}$ is the total length of the carbon fiber rods with diameter $r_\text{c}$ and density $\rho_\text{c}$.

We can write the equation of motion of the wing actuated by the output link of the flapping mechanism as: 
\begin{align*}
	\tau_\text{aero} + \tau_\text{mech} = \tau_\text{inertial}
\end{align*}
where, $\tau_\text{aero}$ is the aerodynamic torque on the wing, $\tau_\text{mech}$ is the torque applied by the mechanism on the wing to execute the required wing kinematics. Thus, the power consumed by the wing motion is given by: 
\begin{align}
	P_\text{mech} 
	= \tau_\text{mech}\, \Omega
	=  \rho\, \Omega_\text{ref}^3 R^3 S \left( \frac{\rho_\text{w}}{\rho}\, \tilde{t}_\text{w}\, \hat{r}_2^3\, \tilde{\Omega}' -  \frac{C_\text{H}}{2} \hat{r}_3^3 \right) \tilde{\Omega}
	\label{eq:power}
\end{align}
Here, $\tilde{t}_\text{w} = t_\text{w}/L_\text{ref}$ is the nondimensional wing--thickness ($L_\text{ref}$: mean chord length), $\hat{r}_3 = r_3/R$ is the nondimensional third moment of wing--area, and $\tilde{\Omega}'$ is the non-dimensional sweeping acceleration.
In the above formulation, we have excluded the inertial power consumed by the flapping mechanism (Sec. \ref{sec:discussion}).

	\section{Results}
\label{sec:results}

We selected three flapping mechanism topologies (Fig. \ref{fig:question}) from the literature to investigate whether optimizing them dimensionally for predefined kinematic parameters such as sweeping amplitudes and sweeping angle profiles is justified. These mechanisms can provide large sweeping amplitudes ($ \ge 120$ deg) without compromising force transmission quality. 
The topologies used by Deng \etal \cite{deng2021design}, Karásek \etal \cite{karásek2014pitch}, and Coleman \etal \cite{coleman2017development} are referred to here as D, K, and C, respectively.  
The wing sweeping velocity profiles were generated using the mechanism simulator and used to evaluate lift and power consumption using 2D-UVLM during the multiobjective optimization, which maximized the lift and minimized the power consumption to yield a Pareto set for each mechanism topology. Finally, we analyzed the Pareto solutions for manufacturing tolerances to select the best solution.

Before optimization, we conducted a convergence study of the 2D-UVLM and validated the method against the experimental data from literature by simulating fruit fly, oval, and figure-8 flapping wing kinematics. To balance computation time with convergence, we chose number of lattice elements $N=20$ and $N_\text{t} = 2$ for all our simulations, where $N_\text{t}$ is defined such that the nondimensional time step $\delta \tilde{t}  = \frac{\tilde{d}/ \tilde{\Omega}_\text{max}}{N_\text{t}}$, and $N_\text{t} = 1$ yields the time for the lattice element to move by one element length $\tilde{d}$ with velocity $\tilde{\Omega}_\text{max}$.  We observed that for Squire's parameter $a=10^{-1}$, convergence was achieved in all cases considered in this study. We simulated for three flapping cycles, and computed the mean of lift and power from the third cycle, as we observed that the lift and drag coefficients are periodic from the third cycle. Details are provided in the supplementary material S1.

\begin{figure*}[!t]
	\centerfloat
	\subfloat[]{\includegraphics[scale=1]{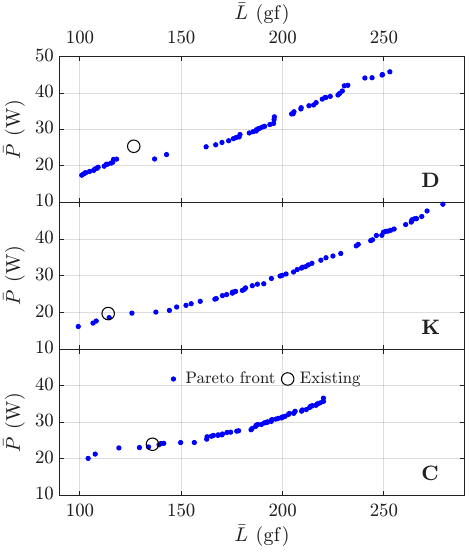}}	
	\hspace*{2em}
	\subfloat[]{\includegraphics[scale=1]{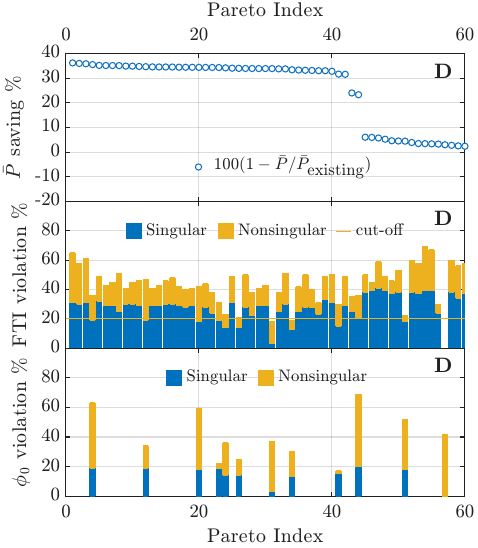}}	
	\vspace*{1em}
	\\
	\subfloat[]{\includegraphics[scale=1]{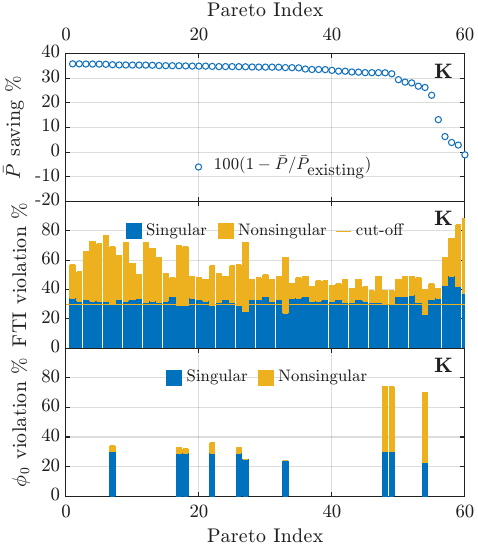}}	
	\hspace*{1em}
	\subfloat[]{\includegraphics[scale=1]{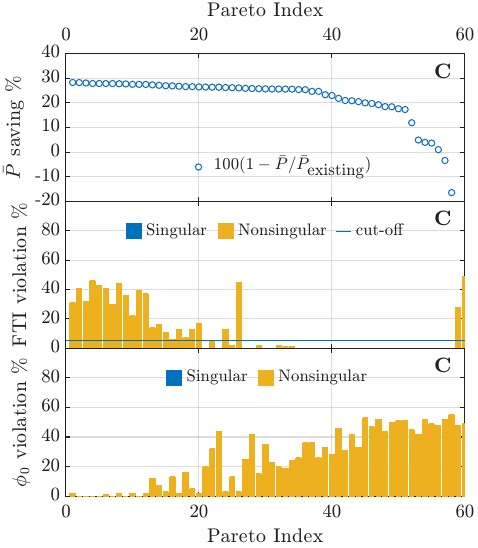}}	
	
	\caption{ Optimization results and robustness analysis:
		\textbf{(a)} Existing mechanisms above Pareto fronts for maximized mean lift ($\bar{L}$) and minimized mean power ($\bar{P}$) for  topologies: D \cite{deng2021design}, K \cite{karásek2014pitch}, C \cite{coleman2017development}.   
		\textbf{Top rows} of \textbf{(b)}--\textbf{(d)}: Significant power savings relative to the existing mechanisms. Top 40 Pareto solutions candidates for the best solution. 
		Manufacturing tolerance analysis: 100 quasi-random linkages simulated in $\pm 0.5$ mm tolerance band of each Pareto solution. \textbf{Middle rows} of \textbf{(b)}--\textbf{(d)}: Significant singularity for D and K for this tolerance.   
		\textbf{Bottom rows} of \textbf{(b)}--\textbf{(d)}: Percentage violation of amplitude constraint, shown for selected Pareto points (singularity \% below the cut-off line shown in middle rows).  
	}
	
	\label{fig:pareto}
\end{figure*}

\begin{figure*}[!t]
	\centering
	\subfloat[]{\includegraphics[scale=1]{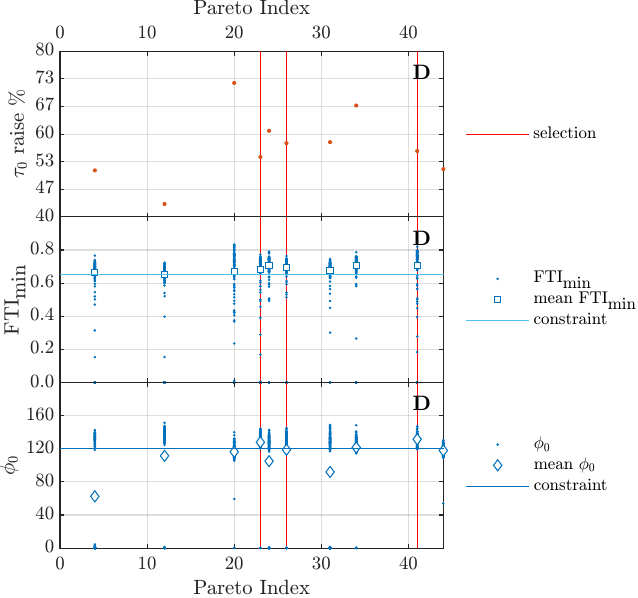}}	
	\vspace*{1em}
	\\
	\subfloat[]{\includegraphics[scale=1]{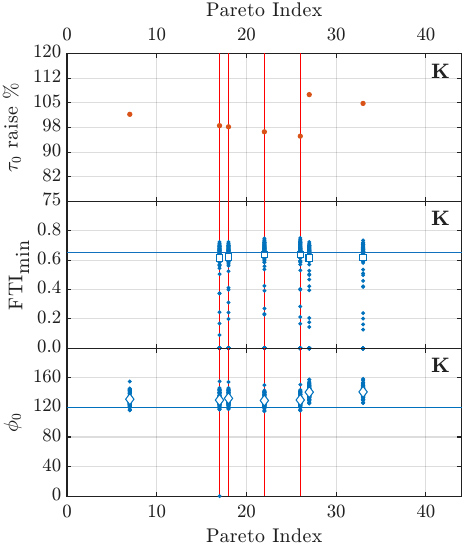}}	
	\hspace*{1em}
	\subfloat[]{\includegraphics[scale=1]{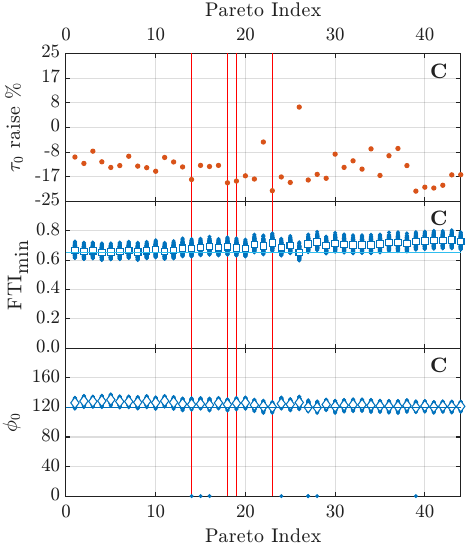}}	
	
	\caption{ Robustness analysis (contd.): \textbf{Top rows} of \textbf{(a)}--\textbf{(c)}: Peak torque increase of selected Pareto solutions relative to the existing mechanisms. Lower the better.
		\textbf{Middle rows} of \textbf{(a)}--\textbf{(c)}: FTI$_\text{min}$ for 100 quasi-random linkages in $\pm 0.5$ mm tolerance band of selected Pareto solutions. Lesser spread and the mean near or above the constraint value - criteria for selected Pareto points (red vertical).   
		\textbf{Bottom rows} of \textbf{(a)}--\textbf{(c)}: Nonsingular linkages with amplitudes $\phi_0$ near zero show defects. Too many defective linkages bring the mean $\phi_0$ down, as in the leftmost case in bottom row of (a). 
	}
	
	\label{fig:paretoSelection}
\end{figure*}

\begin{figure*}[!h]
	\centering
	\subfloat[]{ \includegraphics[scale=1]{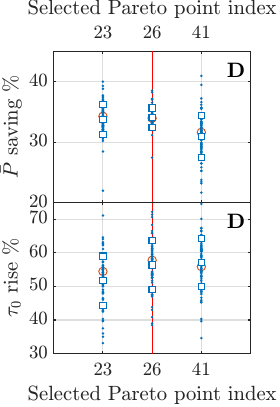}	}
	\subfloat[]{ \includegraphics[scale=1]{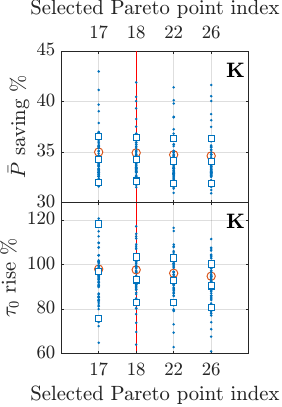}	}
	\subfloat[]{ \includegraphics[scale=1]{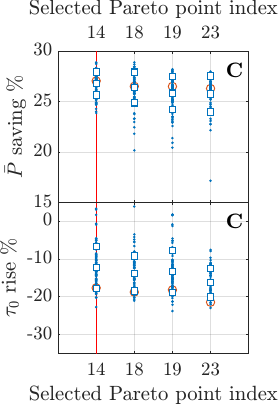}	}				
	\caption{ Robustness analysis (contd.): Mean power savings and peak torque raises in the tolerance band of the selected Pareto solutions in Fig. \ref{fig:paretoSelection} are compared. Their mean and the standard deviation band are marked with a square. The most robust solution (red vertical) is the one with smaller deviation bands while having high mean power saving and low peak torque increase. }
	\label{fig:finalSelection}
\end{figure*}

\begin{figure}[!h]
	\centering
	\includegraphics[scale=1]{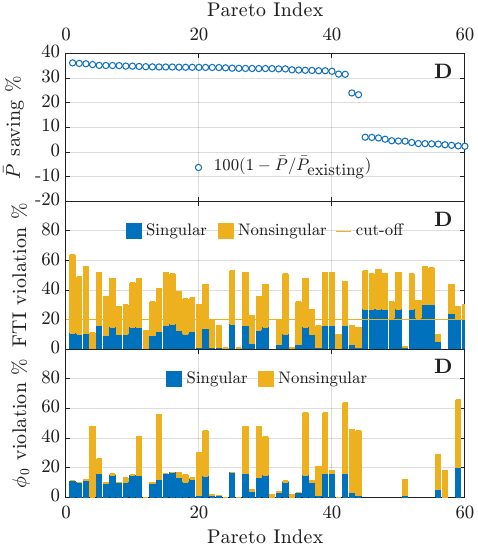}					
	\caption{Robustness analysis for mechanism D for tolerance $\pm 0.15$ mm:  Singularity percentages are significantly reduced compared to tolerance of $\pm 0.5$ mm. FTI$_\text{min}$ violation and $\phi_0$ violations are low for the selected Pareto points indexed 23,26, and 41 in Fig. \ref{fig:paretoSelection}a and negligible for the finally chosen best solution (index-26).}
	\label{fig:higherTolerance}
\end{figure}

\subsection{Optimization constraints and parameters}
\label{sec:optimConstraints}

For the multiobjective optimization, all six constraints were imposed. The minimum amplitude was set to 120 deg. The boxes used for the box constraint are shown in Fig. \ref{fig:question}, and each corresponds to a rectangular boundary that encloses the workspace and the mechanism with the existing dimensions from the literature. The values of $r_\text{max}$ in the constraint \#1 used are 5, 7, and 12 for the mechanisms D, K, and C, respectively, derived from the existing dimensions (literature).
To estimate a value of FTI$_\text{min}$ for the constraint \#5 in Sec. \ref{sec:optimization}, 
we performed a single objective optimization of the mechanism, which maximized the critical FTI$_\text{cr}$ for all three selected mechanisms. The existing mechanism dimensions from the literature were used as the initial guess for each case. The kinematic constraints  \#1--4,6 enlisted in Sec. \ref{sec:optimization} were enforced during the optimization using the patternsearch algorithm in MATLAB. The minimum (= 0.48) of these three critical FTI values was initially used for the FTI constraint for multiobjective optimization. During the multiobjective optimization, we observed that higher critical values are possible for all three mechanisms. We increased FTI$_\text{min}$ to 0.65 for all mechanisms based on those observations. We faced another problem during the multiobjective optimization: it failed to find any feasible solution when we provided the existing mechanism dimensions as initial guesses (which did not satisfy constraints). To address this, we provided the solutions obtained from the single objective optimization as the initial feasible points for searching the Pareto set using the paretosearch() function in MATLAB. 

For the wing pitching velocity, we used the fruit fly kinematics data for all cases (Fig. \ref{fig:optimalKinematics}). Since this work focuses on presenting a tool for optimization of the flapping mechanism (i.e., the sweeping velocity profile), any pitching velocity profile will serve the purpose of demonstration. However, while designing such an MAV, we would need to implement passive wing pitching kinematics in the present framework. 
We assumed $R=22$ cm and $f = 15$ Hz, air density $\rho = 1.225 \times 10^{-3}$ g/cc, kinematic viscosity $\nu = 0.1568$ cm$^2$/s, for all our computations. Further, for the fruit fly wing planform, $r_\text{a} = 3.7425$, $\hat{r}_2^2 = 0.472$, $\hat{r}_3^3 = 0.3415$ (computed from the planform used in robofly experiment \cite{dickinson1999wing}). The choice of these parameters should not affect the comparisons made between the different mechanisms. 

For the finite element simulations, we used $K_\text{in} = 1 $, $K_\text{out} = K_\text{tern} = 10^8$, and $G = 10^9$. The spring attached to the input link is soft. When we lock the mechanism to compute MA, we use a highly stiff spring at the output link. The axial stiffness of truss elements is set to high to simulate their rigid nature. The stiffnesses of springs attached to all other links are set to zero. Note that the units are not mentioned here. Here, only the relative values of these parameters matter because we are concerned only with the nondimensional outputs (output link angles and FTI) from the finite element simulations.

\begin{table*}[!t]
	\centerfloat
	\caption{Selected Pareto solutions: Marked with red vertical lines in Fig. \ref{fig:finalSelection} (a)--(c). 
		Note: $\phi_0$ violation percentages exclude singular linkages from the tolerance band. Such linkages can be defective with two or more partial oscillations of the output for a single crank turn, rendering them unsuitable for flapping (Fig. \ref{fig:defectiveLinkage}). They have close to zero amplitudes in case D (Fig. \ref{fig:paretoSelection}(a)). 
		The nonsingular quasi-random linkages for the selected Pareto solutions were simulated using 2D-UVLM to obtain mean lift and power (Fig. \ref{fig:pareto}(a): gray dots). From that data, mean power decrease \% and peak-input torque increase \% were compared. Their mean and standard deviations are reported here. 
	}
	\label{tab:selectedPareto}
	\begin{tabular}{c c c c c c} 
		\hline 
		\hline 
		\addlinespace[3pt] 
		Mechanism & Tolerance &  Singularity  &   Amplitude 		& $\bar{P}$ 		& $\tau_0$ \\ 
		& (mm) 	  &   \%  		  &   violation \% 		&	decrease \% 	& increase \%  \\ 
		\addlinespace[3pt] 
		\hline 
		\addlinespace[3pt] 
		D  		& $\pm 0.5$	 &  14 	 		&	11		 			& $31 \pm 3$ 	 & $ 56 \pm 7$ \\ 
		K 		& $\pm 0.5$	 &  29 	 		& 	3		 			& $34 \pm 2$ 	 & $ 93 \pm 10 $ \\ 
		C 		& $\pm 0.5$	 &  0 	 		& 	7		 			& $26 \pm 2 $ 	 & $-12\pm 6$ 	\\ 
		
		\hline 
		\hline 
	\end{tabular} 
\end{table*}

\begin{figure*}[t]
	\centerfloat
	\includegraphics[scale=0.9]{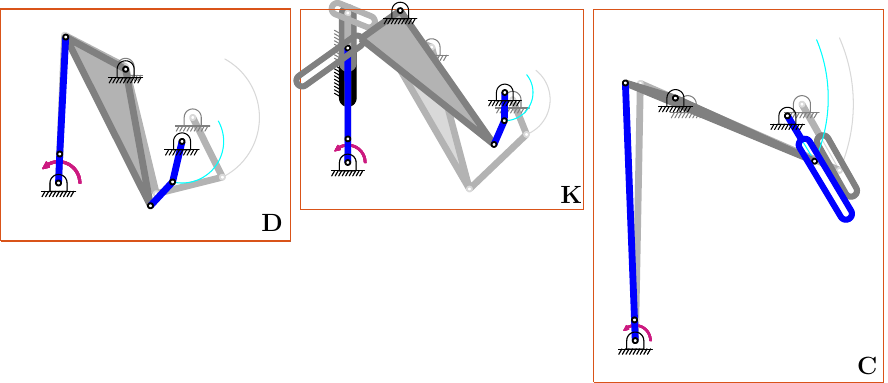}					
	\caption{ Selected optimal mechanisms compared with the existing mechanisms (shown in lighter gray shade): topologies: D - Deng \etal \cite{deng2021design}, K - Karásek \etal \cite{karásek2014pitch}, C - Coleman \etal \cite{coleman2017development}.	}
	\label{fig:optimalMechanisms}
\end{figure*}

\begin{figure}[!h]
	\centering
	\includegraphics[scale=1]{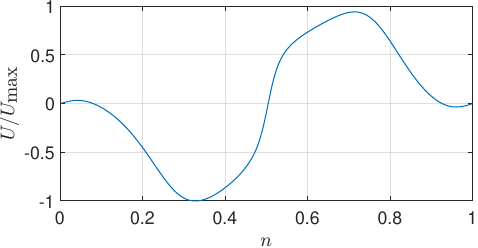}					
	\caption{Defective linkage: The output link oscillates more than once in a flapping cycle ($U=0$ at many points), with very small amplitude in one half-cycle. This is a nonsingular linkage, but undesirable. Such mechanisms violate the amplitude constraint and are eliminated. We defined amplitude as the smaller of the angles traversed in two consecutive half-cycles, when starting with downstroke (Fig. \ref{fig:question}).}
	\label{fig:defectiveLinkage}
\end{figure}

\begin{figure}[!h]
	\centering
	\includegraphics[scale=1]{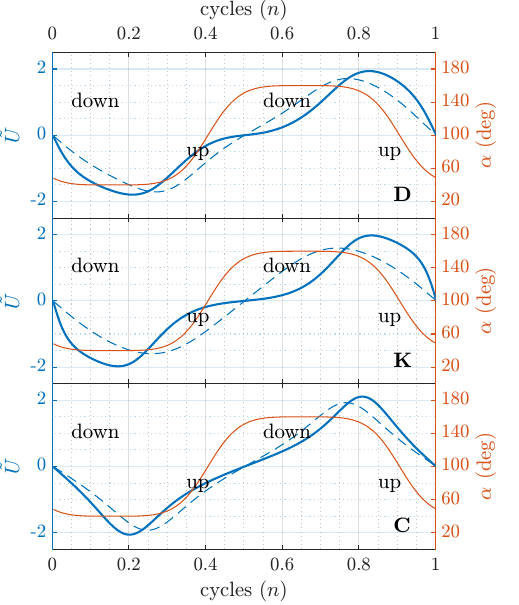}					
	\caption{  Optimal sweeping velocity profiles asymmetric: Nondimensional sweeping velocity $\tilde{U}$ generated by the existing mechanisms in dotted, selected Pareto optimal mechanisms in dark blue: topologies used: D - Deng \etal \cite{deng2021design}, K - Karásek \etal \cite{karásek2014pitch}, C - Coleman \etal \cite{coleman2017development}: Velocities are not symmetric about the midstroke position ($n=0.25, 0.75$). The wing pitching angle $\alpha$ is shown in red. Corresponding lift coefficients are shown in the supplementary material S3. }
	\label{fig:optimalKinematics}
\end{figure}

\begin{table*}[!ht]
	\centerfloat
	\caption{Comparison between existing mechanisms and the selected Pareto solutions listed in Tab. \ref{tab:selectedPareto}: Differences lie in either the sweeping amplitude $\phi_0$ or the mean lift coefficient $\bar{C}_\text{L}$. Flapping frequencies, mean power, and peak-input torque are listed for generating 120 gf of lift from the mechanisms. The selected Pareto solutions consume significantly lower power and operate at lower frequencies.  }
	\label{tab:reasonsComparison}
	
	\begin{tabular}{c @{\hspace{2\tabcolsep}}  @{\hspace{2\tabcolsep}}  c c c c c @{\hspace{2\tabcolsep}}  @{\hspace{2\tabcolsep}} c c c c c  } 
		\hline 
		\hline 
		\addlinespace[3pt] 
		&	\multicolumn{5}{ c @{\hspace{2\tabcolsep}}    @{\hspace{2\tabcolsep}} }{existing}	&	\multicolumn{5}{c}{best} \\ 
		&	\multicolumn{5}{ c @{\hspace{2\tabcolsep}}    @{\hspace{2\tabcolsep}} }{}	& \multicolumn{5}{c}{Pareto point} \\ 
		\addlinespace[3pt] 
		\hline 
		\addlinespace[3pt] 
		& $\bar{C}_\text{L}$ 	 & $\phi_0$  	 & $f$ 	 & $\bar{P}$ 	 & $\tau_0$ 	 & $\bar{C}_\text{L}$ 	 & $\phi_0$  	 & $f$ 	 & $\bar{P}$  	 & $\tau_0$ \\ 
		\addlinespace[3pt] 
		\hline 
		\addlinespace[3pt] 
		units 	 & -- 	 & deg  	 & Hz 	 & W 	 & kg.cm  	 & -- 	 & deg  	 & Hz 	 & W 	 & kg.cm \\ 
		\addlinespace[3pt] 
		D 	 & 1.61 	 & 125 	 & 14.6 	 & 23.5 	 &  8.4 	 & 2.48 	 & 132  	 & 11.1  	 & 15.5 	 & 13.3  \\ 
		K 	 & 1.57 	 & 120 	 & 15.4 	 & 21.3 	 &  6.9 	 & 2.79 	 & 137  	 & 10.1  	 & 13.9 	 & 13.7  \\ 
		C 	 & 1.84 	 & 121 	 & 14.1 	 & 19.9 	 &  8.8 	 & 2.58 	 & 125  	 & 11.5  	 & 14.5 	 &  7.3  \\ 
		\addlinespace[3pt] 
		\hline 
		\hline 
	\end{tabular} 
	
\end{table*}

\subsection{Optimization results}
\label{sec:results_pareto}

The Pareto fronts for the three mechanism topologies were obtained for a flapping frequency of 15 Hz (Fig. \ref{fig:pareto}(a): blue dots).  We observed that when we altered the frequency to meet a specific lift requirement, the mean power consumption did not vary much for 40 out of 60 Pareto points (Fig. \ref{fig:pareto}(b), (c),(d): top rows). Therefore, we considered additional criteria for picking one for an MAV design rather than just choosing the one with the highest power savings. The mean power consumption and peak torque requirements are essential to select actuators such as brushless motors with added gearboxes. We must ensure that the selected mechanism leads to high mean power savings and low peak-input torque raises. The manufacturing tolerances could also adversely impact the mechanism's performance. Therefore, we conducted a robustness analysis of the mechanisms.

Additive manufacturing, such as fused deposition molding (FDM), is a common method for 3D printing of flapping wing MAV parts. FDM's tolerance ranges between $\pm 0.2$ and $\pm 0.5$ mm. We chose the upper limit $\pm 0.5$ mm to analyze the uncertainty in link--lengths from FDM. Assuming a crank--length of 10 mm for all mechanisms, we generated a set of 100 linkages from a selected Pareto solution using 100 quasi-random points ($\vect{x}$ in Eqs. \ref{eq:vectorSobol}: $l$ and $\angle$ denote respectively, the link length and included angle between respective nodes numbered in subscripts: Fig. \ref{fig:question}) using the Sobol sequence \cite{cao2016application}. The Sobol sequence is regarded as the best method to generate quasi-random numbers, which perform significantly better than pseudo-random numbers in uncertainty analysis due to faster convergence and even coverage of the domain space. 

\begin{align}
	\text{D}: \vect{x} &= \{l_{12}, l_{23}, l_{34}, l_{41}, l_{45}, l_{56}, l_{67}, l_{74}, l_{71}, \angle_{345}\} \nonumber \\
	\text{K}: \vect{x} &= \{l_{12}, l_{23}, x_{3}, x_4, l_{41}, l_{45}, l_{56}, l_{67}, l_{74}, l_{71}, \angle_{345}\} \nonumber\\
	\text{C}: \vect{x} &= \{l_{12}, l_{23}, l_{34}, l_{41}, l_{45}, l_{64}, l_{61}, \angle_{345}\}  
	\label{eq:vectorSobol}
\end{align}

Post-generation, we analyzed the linkages in the tolerance band for the critical constraint violations -- quick return ratio, Grashof's condition, amplitude, and FTI (singularity). The box constraints were not violated much to cause concern due to small link-length changes. The quick return ratio (QRR) remained in the $1 \pm 0.02$ range for all three mechanisms, D, K, and C, in their tolerance bands. Grashoff's condition was satisfied by all 100 linkages in the bands for each of the three mechanisms due to the offset value 0.4 used in the optimization. The amplitude $\phi_0$ and FTI$_\text{min}$ values for the quasi-random linkages in the tolerance bands of all Pareto solutions are shown in Fig. \ref{fig:pareto} (b), (c), and (d) (middle and bottom rows). We found that the singularity percentage, i.e., the number of quasi-random linkages having singular configurations out of the 100, hovered between 10 and 50 for the mechanisms D and K, and zero for the mechanism C. The number of nonsingular quasi-random linkages violating the FTI$_\text{min}$ constraint lay between 5 and 50 for mechanisms D and K, and 0 and 50 for mechanism C. Further, the number of nonsingular quasi-random linkages violating the amplitude constraint (<120 deg) lay between 0 and 50 for all three mechanisms. Given these high values, discarded Pareto solutions with singularity percentage above a cut-off value( middle rows: Fig. \ref{fig:pareto}b--d) to get a smaller set (bottom rows: Fig. \ref{fig:pareto}b--d). In this set, we compared the spread in FTI$_\textrm{min}$ and $\phi_0$ and their mean (Fig. \ref{fig:paretoSelection}). We discarded those with the mean values significantly lower than the constraint values. Final selections were made based on mean power and peak-input torque values (Fig. \ref{fig:finalSelection}). As a result, we found that the selected Pareto solution has a small variation in mean power savings for a given lift (Tab. \ref{tab:selectedPareto}) among the feasible quasi-random linkages. The variation in peak-input torque is larger. This, and the fact that many mechanisms in the tolerance band are singular, indicate that the chosen tolerance value of $\pm 0.5$ mm is large, and we must employ other fabrication methods that can offer high manufacturing tolerance. We simulated the selected Pareto solution for mechanism D in the $\pm 0.15$ mm tolerance band. The constraint violation and singularity percentage dropped significantly (Fig. \ref{fig:higherTolerance}, Tab. \ref{tab:selectedPareto}: row 2). The ranges of mean power savings and peak-input torque increase were significantly lowered, thus increasing the reliability of the prototype upon fabrication. 

The selected Pareto optimal flapping mechanisms are shown in Fig. \ref{fig:optimalMechanisms} overlapped with the existing mechanisms in the literature.
Comparison of the existing mechanisms with the selected Pareto solutions shows that the optimized ones have amplitudes slightly higher, lift-coefficients significantly higher, and need lower frequencies to generate the same lift (Tab. \ref{tab:reasonsComparison}), consuming 26--34\% less power (Tab. \ref{tab:selectedPareto}), but requiring higher peak-input torques. 
The sweeping velocity profiles for the selected Pareto solutions are all asymmetric, unlike those for the existing mechanisms (Fig. \ref{fig:optimalKinematics}).

Finally, Tabs. \ref{tab:selectedPareto} and \ref{tab:reasonsComparison} reveal that the topology C  (i) is least sensitive to manufacturing tolerances as it has very few singular mechanisms, and very small amplitude variations in the tolerance band, and (ii) requires significantly lower peak torque compared to the other two, for the same lift, at nearly similar power. 

	\section{Discussion}
\label{sec:discussion}

Rigid link flapping mechanisms are still the most practical choice for flapping wing MAVs to carry a useful payload and an onboard battery for free flight due to their long-term durability and reliability. While designing a flapping wing MAV with such mechanisms for high agility, maneuverability, and hovering ability like insects, it is crucial to reduce its weight as much as possible. One way to achieve that is by using single-DOF planar rigid linkages to generate the sweeping component of the flapping motion and optimizing its amplitude and the output link velocity profile for high lift and low power, such that smaller (and lesser) motors and batteries can be used. Existing rigid link flapping mechanisms are rarely optimized by computing lift and power in situ, and rather focus on replicating predefined amplitude and sinusoidal velocity profiles, both of which need not be optimal, as shown in this study. The suboptimal selection of components for these MAVs is evident from short flight durations and motor failure due to overheating. The present work addresses these issues by combining a mechanism simulator with an aerodynamic analysis tool in a single optimization routine, thereby (i) optimizing the flapping mechanisms and (ii) providing a computational tool for the optimal selection of the MAV components. The mechanism simulator can analyze any generic planar single-DOF rigid link mechanism except for inclined sliders.

\subsection{Optimization of flapping mechanisms}

We optimized one-DOF planar rigid-link flapping mechanisms for high mean lift and low mean power consumption for a given flapping frequency and wing morphology. 
We observed from the study of three mechanism topologies that following the above procedure can achieve significant power saving (26--34\%). We also saw that improvements in mechanisms are not necessarily from an increase in flapping amplitude but also from improved sweeping velocity profiles generated by the mechanisms. The asymmetric sweeping velocity profiles from the optimal mechanisms show that assuming a symmetric velocity profile a priori is not justified. We also performed uncertainty analysis due to manufacturing tolerances, revealing that 3D printing using FDM may not be a suitable method for fabrication due to the low tolerances available ($\pm0.5$ mm). With that, there is a significantly high chance of failure due to singularity or amplitude defects. Further, even for the best Pareto solution, the tolerance band's worst-case would result in 7\% more power consumption than the corresponding Pareto solution for the same lift (Fig. \ref{fig:finalSelection}, mechanism D). When we reduced tolerance in the range of CNC machining ($\pm0.15$ mm), the chances of failure due to singularity or amplitude defects reduced significantly, and became negligible for the selected solution.
Therefore, higher tolerances are essential in manufacturing such mechanisms to obtain reliable designs. Further, one should be careful in material selection to include the effects of wear and tear and operating temperatures on dimensional alterations. 

We observed that different mechanism topologies require varied power and peak input torques for a given design mean lift value. Thus, the presented method helps select the best mechanism topology while designing an MAV. Further, it may be combined with an optimization routine to obtain an optimal topology for the flapping mechanism since the nonlinear finite element model can be used to analyze any generic planar rigid link mechanism of one DOF with binary and ternary links. 

In the current approach, we fixed the flapping frequency $f$, wing size $R$, wing shape (fruit fly), and pitching angle profile $\alpha(t)$ and optimized the flapping mechanism dimensions. Wing shape and size are independent of the mechanism. Therefore, fixing them beforehand is justified. We assume $\alpha(t)$ to demonstrate the usefulness of the optimization. Practically, such mechanisms are used with passive wing pitching kinematics, which means $\alpha(t)$ during stroke reversal changes automatically due to the torsional stiffness of the joint of the wing-root with the fuselage, and aerodynamic torques. During the translation phase, the angle can be kept constant through mechanical stoppers. Thus, $\alpha(t)$ can be similar to the fruit fly case. At this point, one might inquire whether fixing the flapping frequency $f$ a priori is justified. During the optimization, the lift $\bar{L}$ and power $\bar{P}$ vary due to changes in the sweeping amplitude $\phi_0$ and the lift coefficient $C_\text{L}$, due to changes in the link lengths. 
While varying $f$  in a small range (10-20 Hz), we assumed the mean lift coefficient constant. This assumption is based on the observation that $\bar{C}_\text{L}$ varies very little in this frequency range (supplementary material S2). With this assumption, we can say from Eqs. \eqref{eq:lift} and \eqref{eq:power} that: 
\begin{align}
	& \bar{L} = c f^2, & \bar{P} = d f^3 
\end{align}
where, $c$ and $d$ are proportionality constants independent of $f$.
Therefore, compared to a reference mechanism (say, the existing one) with lift and power values $\bar{L}_\text{r}$ and $\bar{P}_\text{r}$ respectively at frequency $f_\text{r}$, a Pareto solution with the same lift $\bar{L}_\text{r}$ at a different frequency $f$, will consume power $\bar{P}$ given by: 
\begin{align}
	\frac{\bar{P}}{\bar{P}_\text{r}} = \frac{d}{d_\text{r}} \left(\frac{c_\text{r}}{c} \right)^{3/2}
\end{align}
$c$, $d$, $c_\text{r}$, $d_\text{r}$ are the respectively proportionality constants. The above equation implies that the power ratio is independent of the flapping frequency $f_\text{r}$, and thus has the same value for any lift $L_\text{r}$. Thus, provided all mechanisms generate the same lift, a mechanism that is $x\%$ better than the reference mechanism for that lift will always be $x\%$ better for every other lift. It means the mechanism that consumes the least power for that lift will consume the least power for every other lift $L_\text{r}$, i.e., for every other frequency $f_\text{r}$. Therefore, allowing the lift to vary in the multiobjective optimization while keeping $f$ unchanged is equivalent to having a constant lift and allowing $f$ to vary freely. It might suggest using a single objective optimization with mean power minimization for a constant lift. However, we observed from our robustness analysis that the mechanism that consumes the least power may not be the most reliable and robust when manufactured. A compromise between the ideal best performance and reliability is required. To realize this, we need multiple competing solutions beforehand. Further, one may also have different requirements, such as a high-performance MAV at the expense of less energy efficiency. The multiobjective optimization provides such options at one's disposal.

\subsection{Optimal selection of components for MAV design}

Among MAVs with large amplitude 1-DOF rigid-link planar flapping mechanisms, only the Colibri by Roshanbin \etal \cite{roshanbin2017colibri}, based on the mechanism by Karásek \etal \cite{karásek2014pitch}, and the Robotic Hummingbird by Coleman \etal  \cite{coleman2017development} were tested for free flight. Colibri with a 160 mAh battery sustained flight for 15-20 s, while Robotic Hummingbird with a 180 mAh battery lasted for 50-60 s until the motors overheated. These flight times are insufficient for practical purposes. Bigger, heavier batteries need to be incorporated for longer runtime. However, the increased weight implies that we must reconsider many design aspects, such as the wing size, flapping frequency, and motor power rating. An improperly chosen motor cannot supply the required power without overheating or can be overpowered and heavy, rendering the design suboptimal. Since there are many parameters to consider, it is essential to have a computational tool to help find their optimal combination. The tool would reduce trial and error and experiments. The information from the present analysis can help in this regard.

Given the payload and flight time requirements, we can perform a preliminary selection of motors, batteries, wing size, and flapping frequency and estimate the overall weight and lift requirement. We can compute the power and peak-input torque requirements for this lift for the selected mechanism. If the selected motors are overpowered or underpowered, we iterate by choosing a different motor. If the estimated flight time from the power requirement is insufficient, we choose a bigger battery. Then, we have a revised weight estimate and lift requirement estimate. We reiterate this selection procedure until the requirements are met. The peak-input torque information also helps choose an appropriate gear ratio for the selected motor. In Tab. \ref{tab:reasonsComparison}, we notice that the peak input torque increases in the case of the selected Pareto solution for all three mechanism topologies. It implies that a higher gear ratio would be needed in such a case. This increased gear ratio would result in some power loss due to gear friction and noise. These losses are not accounted for in this study.  

\subsection{Comments on choices and assumptions}

In this analysis, we have excluded the inertial power consumption due to the mechanism for which a detailed CAD design would be required. The design must also be structurally optimized for deflection, fatigue, and stress failures. To include a crude estimate, we may increase the lift requirement by a factor of safety. Deng \etal \cite{deng2021design} noticed during their experiments that the mechanism alone consumes 0.13 W at 20 Hz when used without the wing onboard. However, to generate 27.8 gf of lift, the power consumption increased to 9.9 W. Therefore, the inertial power consumed by the mechanism is only a tiny fraction of the total power consumed. The above two reasons are sufficient to justify the exclusion of the inertial power computation of the mechanism.   

In our analysis, we assumed both the revolute and sliding joints to be frictionless. If not properly lubricated, friction in sliding joints is expected to increase power consumption considerably compared to the revolute joints. Therefore, incorporating them would improve the accuracy of our analysis further. Alternatively, revolute joint-only designs involving miniature bearings should be preferred, resulting in low friction. 

We used a two-dimensional version of the extended UVLM for our computations to leverage its computational speed. We also implemented the 3D extended UVLM and compared its performance with the 2D-UVLM (supplement \ref{sec:app:2d_3d}). We observed that both versions captured the lift and drag trends well. Some peaks and valleys in 2D-UVLM were overestimated, and some were underestimated in 3D-UVLM. The mean lift and drag coefficients show similar deviations from the experimental values, indicating no clear advantage of 3D-UVLM. However, the 2D-UVLM performed about 5 times faster than the 3D-UVLM - providing a significant advantage in optimizations where many function evaluations are required.

\section{Conclusion}
\label{sec:conclusion}

A unified computational tool was developed that combines a generic planar single-DOF rigid link mechanism simulator with an unsteady vortex lattice aerodynamic model for multiobjective optimization of flapping mechanisms. We achieved significant power savings (26--34\%), due to larger amplitudes and increased mean lift coefficient resulting from asymmetry in sweeping velocity profiles. The robustness analysis quantified the performance sensitivity to small variations in link dimensions that may result from manufacturing tolerances, assembly misalignments, or aging. The Pareto set and robustness analysis helped select the best mechanism dimensions through a trade-off between performance and reliability. Finally, the procedure helped select the best mechanism topology, as we observed significant variation in sensitivity to manufacturing tolerances and peak input torque values across different topologies for a given design lift value. 

The quantification of power, peak input torque, and lift values can be helpful in optimally selecting actuators, gear ratios, and battery capacity to meet the required payload and flight time requirements from a flapping wing MAV with a given mechanism topology. This computational tool can also be incorporated in a flapping mechanism topology optimization routine, as it can simulate any generic single-DOF planar rigid link mechanism without supplying the mechanism kinematics manually. 
	
		\section*{Acknowledgment}		
		 This work was supported by Visvesvaraya PhD Scheme, MEITY, Government 
	Of India MEITY-PHD-2065.
		
	\appendix
	
		\appendix

\section{2D-UVLM}
\label{app:uvlm}

In hovering flight, for a finite wing sweeping with angular velocity $\Omega$, the translation speed at radial distance $r$ from the rotation axis is $U = \Omega r$. 
The equations given here are non-dimensionalized to eliminate the effect of wing size and frequency. $\tilde{a}$ denotes nondimensional $a$. When using the blade element theory, we use the following reference scales for a wing chord of length $c$ at the radial distance $r$ from the rotation axis:
\begin{align*}
	L_\text{ref} &= c, &
	\Omega_\text{ref} &= 2\phi_0\, f, &
	U_\text{ref} &= \Omega_\text{ref}\, r, \nonumber \\ 
	T_\text{ref} &= \frac{L_\text{ref}}{U_\text{ref}}, & 
	\dot{\alpha}_\text{ref} &= \frac{1}{T_\text{ref}}
\end{align*}	
where, $\phi_0$ is the sweeping (flapping) amplitude, and $f$ is the flapping frequency. We use the mean sweeping angular velocity as the scale $\Omega_\text{ref}$ for sweeping angular velocity. For wing pitching velocity $-\dot{\alpha}(t)$, we use  $1/T_\text{ref}$ (can be verified from Eq. \eqref{eq:impenetrability}) as the reference scale $\dot{\alpha}_\text{ref}$. As a consequence of the blade element theory,  $r=r_2$ results in a suitable velocity reference scale $U_\text{ref} = \Omega_\text{ref} r_2$  for the entire wing, where $r_2$ is the second moment of wing area. In the case of flapping translation velocity in hovering, for different wing-sizes, force coefficients converge better \cite{lua2014scaling} with this velocity reference $U_\text{ref} = 2\phi_0\, f\, r_2$, than with wing-tip velocity $2\phi_0\, f\, R$.

The impenetrability  condition at the  collocation point $\text{CP}_j$ on $j^\text{th}$ element ($j=1,2,\cdots, N$):
\begin{align}
	\underbrace{( \tilde{\vect{u}}^\text{b}_j + \tilde{\vect{u}}^\text{w}_j ) \cdot \uvect{n}_j }_\text{\parbox{2.3cm}{flow velocity}} = 
	\underbrace{ \left[\tilde{\Omega} \unitvector{s} -  \tilde{\dot{\alpha}} \uvect{k} \times ( \tilde{\vect{r}}_j^\text{CP} -\tilde{\vect{r}}_\text{O})\right] \cdot \uvect{n}_j
	}_\text{\parbox{3cm}{chord-element velocity}}
	\label{eq:impenetrability}
\end{align}
where, $\tilde{\vect{u}}^\text{b}_j$ and $\tilde{\vect{u}}^\text{w}_j$ are the flow velocities due to all $N+1$ bound and all $N_\text{w}$ wake vortices at the collocation point CP$_j$ with position $\tilde{\vect{r}}_j^\text{CP}$, given by Eqs. \eqref{eq:bound_velocity}--\eqref{eq:wake_velocity}. $\unitvector{n}_j$ is the normal to the wing chord at that point. $\uvect{k}$ is the normal to the plane $x$--$y$ of the wing-chord. $\unitvector{s}$ indicates the direction of the velocity $U$.
This condition yields $N$ equations in $N+1$ unknowns $\Gamma^\text{b}_i$, $i=1,2,3, \cdots, (N+1)$.
\begin{align}
	\tilde{\vect{u}}^\text{b}_j &= \sum_{i=1}^{N+1} \frac{\tilde{\Gamma}_i^\text{b}}{2\pi} \frac{\uvect{k} \times (\tilde{\vect{r}}_j^\text{CP} -\tilde{\vect{r}}_i^\text{b})  }{ \|\tilde{\vect{r}}_j^\text{CP} -\tilde{\vect{r}}_i^\text{b} \|^2  }	
	\label{eq:bound_velocity}
	\\
	\tilde{\vect{u}}^\text{w}_j &= \sum_{i=1}^{N_\text{w}} \frac{\tilde{\Gamma}_i^\text{w} }{2\pi} \frac{\uvect{k} \times (\tilde{\vect{r}}_j^\text{CP} -\tilde{\vect{r}}_i^\text{w}) }{ \sqrt{ \|\tilde{\vect{r}}_j^\text{CP} -\tilde{\vect{r}}_i^\text{w} \|^4  + \tilde{r}_\text{c}^4 } }	
	\label{eq:wake_velocity}
\end{align}
Here, $\tilde{\vect{r}}_i^\text{b}$ and $\tilde{\vect{r}}_i^\text{w}$ are positions of bound vortices $\tilde{\Gamma}_i^\text{b}$ and wake vortices $\tilde{\Gamma}_i^\text{w}$ respectively. To model the viscous diffusion in wake vortices, a vortex-core growth model proposed by Ramasamy \etal  \cite{ramasamy2007reynolds}, which considers eddy viscosity and improves over the Lamb Oseen model, is used. We used Vatistas' model (n=2) for velocity profile instead of the Lamb Oseen for faster computations, while being very close to the Lamb Oseen profile \cite{vatistas1991simpler}. The nondimensional core radius $\tilde{r}_\text{c}$  is given by:
\begin{align*}
	\tilde{r}_\text{c}^2 &= 4\alpha_\text{L} \left( \frac{1}{Re} + a_1 \tilde{\Gamma}_i^\text{w}   \right)  \tilde{t}
\end{align*}
where, $\alpha_\text{L}$ is the Lamb constant, $a_1$ is the Squire's parameter, $\nu$ is the kinematic viscosity of the medium, and $\tilde{t}$ is the nondimensional time since vortex shedding. Reynolds number in hovering flight 
\begin{align*}
	Re = \frac{U_\text{ref}L_\text{ref}}{\nu} = (2\phi_0 f\, R^2) \,  \frac{\hat{r}_2}{r_\text{a}\, \nu} 
\end{align*}
where, $\hat{r}_2$ is the nondimensional second moment of the wing area, $R$ is the distance of the wing-tip to the axis of wing revolution, $r_\text{a}$ is the wing aspect ratio $=R/\bar{c}$, $\bar{c}$ being the mean chord length, $\nu$ the kinematic viscosity. 

The Kelvin circulation theorem yields the $(N+1)^\text{th}$ equation
\begin{align*}
	\sum_{j=1}^{N+1} \tilde{\Gamma}_j^\text{b} + \sum_{j=1}^{N_\text{w}} \tilde{\Gamma}_j^\text{w} = 0 
\end{align*}
The $N+1$ equations are linear in $N+1$ unknowns $\Gamma^{b}_j$ and can be solved easily.

Using the unsteady Bernoulli equation in terms of the moving frame $x$--$y$ \cite{milnethomson}\footnote{Sec. 3.61 of the book} attached to a thin flat wing-chord, the pressure coefficient difference $\Delta C_\text{p} = (p_1 - p_2)/(\frac{1}{2}\rho U_\text{ref}^2) $ at $k^\text{th}$ time step $t_k$, for $j^\text{th}$ element, can be computed at the collocation point $\text{CP}_j$ as: 
\begin{align}
	\Delta C_\text{p} = 2 \bigg[ \left(- \tilde{\Omega} \, \unitvector{s} \cdot \uvect{x} + \tilde{\vect{u}}^\text{w}_j \cdot \uvect{x} \right) \frac{\tilde{\Gamma}_{j,k}^\text{b}}{\tilde{d}} 
	\nonumber \\ 
	+  \sum_{i=1}^j  \frac{\tilde{\Gamma}_{i,k}^\text{b} -  \tilde{\Gamma}_{i,k-1}^\text{b}}{\tilde{t}_k - \tilde{t}_{k-1}} \bigg]
	\label{eq:nondim_deltaPressureCoeff}
\end{align}
where, $\tilde{d}$ is element length, $\tilde{\Gamma}^b_{j,k}$ is the circulation of $j^\text{th}$ bound vortex element at time $\tilde{t}_k$. 
Sides 1 and 2 of the chord are shown in Fig. \ref{fig:aerodynamic}c; 2 corresponds to the negative $y$ half-plane. 

The pressure-force coefficient $\mathbf{C}_\text{F}$ on the wing-chord, with magnitude $C_\text{F}$: 
\begin{align}
	\mathbf{C}_\text{F}  &=  - \uvect{y}  \sum_{j=1}^N \Delta C_\text{p} \, \tilde{d}  \nonumber\\
	&=  C_\text{F} \big( -\sin\alpha \, \uvect{\bm{\phi}} + \cos\alpha \, \uvect{Z} \big)
	\label{eq:force_ceoff}
\end{align}
Thus, lift coefficient i.e., the vertical component: 
\begin{align*}
	C_\text{L} = C_\text{F}  \cos\alpha
\end{align*}
And, the horizontal component:
\begin{align*}
	C_\text{H} = -C_\text{F} \sin\alpha 
\end{align*}
Conventionally, drag is the force opposite to the motion. Hence, the drag coefficient is: 
\begin{align*}
	C_\text{D} = -\text{sign}(U)\, C_\text{H}
\end{align*}


	\bibliographystyle{asmejour}
	{\footnotesize
	\bibliography{references}}

\begin{thebibliography}{10}
\newcommand{\enquote}[1]{``#1''}
\providecommand{\url}[1]{\texttt{#1}}
\providecommand{\urlprefix}{}
\expandafter\ifx\csname urlstyle\endcsname\relax
  \providecommand{\doi}[1]{doi:\discretionary{}{}{}#1}\else
  \providecommand{\doi}{doi:\discretionary{}{}{}\begingroup
  \urlstyle{rm}\Url}\fi
\providecommand{\eprint}[2][]{\urlprefix\url{#1#2}}
\providecommand{\hrefurl}[2]{\href{#1}{#2}}

\bibitem{song2023review}
Song, F., Yan, Y., and Sun, J., 2023, \enquote{Review of insect-inspired wing
  micro air vehicle,}
  \hrefurl{https://doi.org/10.1016/J.ASD.2022.101225}{Arthropod Structure and
  Development}, \textbf{72}, p. 101225.

\bibitem{singh2022classification}
Singh, S., Zuber, M., Hamidon, M.~N., Mazlan, N., Basri, A.~A., and Ahmad,
  K.~A., 2022, \enquote{Classification of actuation mechanism designs with
  structural block diagrams for flapping-wing drones: A comprehensive review,}
  \hrefurl{https://doi.org/10.1016/j.paerosci.2022.100833}{Progress in
  Aerospace Sciences}, \textbf{132}.

\bibitem{yousaf2021recent}
Yousaf, R., Shahzad, A., Qadri, M.~M., and Javed, A., 2021, \enquote{Recent
  advancements in flapping mechanism and wing design of micro aerial vehicles,}
  \hrefurl{https://doi.org/10.1177/0954406220960783}{Proceedings of the
  Institution of Mechanical Engineers, Part C: Journal of Mechanical
  Engineering Science}, \textbf{235}, pp. 4425--4446.

\bibitem{bergou2007passive}
Bergou, A.~J., Xu, S., and Wang, Z.~J., 2007, \enquote{Passive wing pitch
  reversal in insect flight,}
  \hrefurl{https://doi.org/10.1017/S0022112007008440}{Journal of Fluid
  Mechanics}, \textbf{591}, pp. 321--337.

\bibitem{whitney2010aeromechanics}
Whitney, J.~P. and Wood, R.~J., 2010, \enquote{Aeromechanics of passive
  rotation in flapping flight,}
  \hrefurl{https://doi.org/10.1017/S002211201000265X}{Journal of Fluid
  Mechanics}, \textbf{660}, pp. 197--220.

\bibitem{hu2019effects}
Hu, F. and Liu, X., 2019, \enquote{Effects of stroke deviation on hovering
  aerodynamic performance of flapping wings,}
  \hrefurl{https://doi.org/10.1063/1.5124916/988366}{Physics of Fluids},
  \textbf{31}, p. 111901.

\bibitem{luo2018effects}
Luo, G., Du, G., and Sun, M., 2018, \enquote{Effects of stroke deviation on
  aerodynamic force production of a flapping wing,}
  \hrefurl{https://doi.org/10.2514/1.J055739}{AIAA Journal}, \textbf{56}, pp.
  25--35.

\bibitem{sane2001control}
Sane, S.~P. and Dickinson, M.~H., 2001, \enquote{The control of flight force by
  a flapping wing: lift and drag production,}
  \hrefurl{https://doi.org/10.1242/JEB.204.15.2607}{Journal of Experimental
  Biology}, \textbf{204}, pp. 2607--2626.

\bibitem{wang2016numerical}
Wang, C., Zhou, C., and Xie, P., 2016, \enquote{Numerical investigation into
  the effects of stroke trajectory on the aerodynamic performance of insect
  hovering flight,} \hrefurl{https://doi.org/10.1007/S12206-016-0322-3}{Journal
  of Mechanical Science and Technology}, \textbf{30}, pp. 1659--1669.

\bibitem{zhang2016optimization}
Zhang, H., Wen, C., and Yang, A., 2016, \enquote{Optimization of lift force for
  a bio-inspired flapping wing model in hovering flight,}
  \hrefurl{https://doi.org/10.1177/1756829316653698}{International Journal of
  Micro Air Vehicles}, \textbf{8}, pp. 92--108.

\bibitem{phan2019extremely}
Phan, H.~V., Truong, Q.~T., and Park, H.~C., 2019, \enquote{Extremely large
  sweep amplitude enables high wing loading in giant hovering insects,}
  \hrefurl{https://doi.org/10.1088/1748-3190/AB3D55}{Bioinspiration and
  Biomimetics}, \textbf{14}, p. 066006.

\bibitem{deng2021design}
Deng, H., Xiao, S., Huang, B., Yang, L., Xiang, X., and Ding, X., 2021,
  \enquote{Design optimization and experimental study of a novel mechanism for
  a hover-able bionic flapping-wing micro air vehicle,}
  \hrefurl{https://doi.org/10.1088/1748-3190/abc292}{Bioinspiration and
  Biomimetics}, \textbf{16}.

\bibitem{karásek2014pitch}
Karásek, M., Hua, A., Nan, Y., Lalami, M., and Preumont, A., 2014,
  \enquote{Pitch and roll control mechanism for a hovering flapping wing MAV,}
  \hrefurl{https://doi.org/10.1260/1756-8293.6.4.253}{International Journal of
  Micro Air Vehicles}, \textbf{6}, pp. 253--264.

\bibitem{coleman2017development}
Coleman, D., Benedict, M., Hirishikeshaven, V., and Chopra, I., 2017,
  \enquote{Development of a robotic hummingbird capable of controlled hover,}
  \hrefurl{https://doi.org/10.4050/JAHS.62.032003}{Journal of the American
  Helicopter Society}, \textbf{62}.

\bibitem{jeon2017design}
Jeon, J.~H., Cho, H., Kim, Y., Lee, J.~H., Gong, D.~H., Shin, S.~J., and Kim,
  C., 2017, \enquote{Design and analysis of the link mechanism for the flapping
  wing MAV using flexible multi-body dynamic analysis,}
  \hrefurl{https://doi.org/10.1177/1756829316682148}{International Journal of
  Micro Air Vehicles}, \textbf{9}, pp. 253--269.

\bibitem{lin2002force}
Lin, C.-C. and Chang, W.-T., 2002, \enquote{The force transmissivity index of
  planar linkage mechanisms,}
  \hrefurl{https://doi.org/10.1016/S0094-114X(02)00070-8}{Mechanism and Machine
  Theory}, \textbf{37}(12), pp. 1465--1485.

\bibitem{berman2007energy}
Berman, G.~J. and Wang, Z.~J., 2007, \enquote{Energy-minimizing kinematics in
  hovering insect flight,}
  \hrefurl{https://doi.org/10.1017/S0022112007006209}{Journal of Fluid
  Mechanics}, \textbf{582}, pp. 153--168.

\bibitem{nabawy2015aero}
Nabawy, M.~R. and Crowther, W.~J., 2015, \enquote{Aero-optimum hovering
  kinematics,}
  \hrefurl{https://doi.org/10.1088/1748-3190/10/4/044002}{Bioinspiration and
  Biomimetics}, \textbf{10}, p. 044002.

\bibitem{yan2015effects}
Yan, Z., Taha, H.~E., and Hajj, M.~R., 2015, \enquote{Effects of aerodynamic
  modeling on the optimal wing kinematics for hovering MAVs,}
  \hrefurl{https://doi.org/10.1016/J.AST.2015.04.013}{Aerospace Science and
  Technology}, \textbf{45}, pp. 39--49.

\bibitem{ke2017wing}
Ke, X., Zhang, W., Cai, X., and Chen, W., 2017, \enquote{Wing geometry and
  kinematic parameters optimization of flapping wing hovering flight for
  minimum energy,}
  \hrefurl{https://doi.org/10.1016/j.ast.2017.01.019}{Aerospace Science and
  Technology}, \textbf{64}, pp. 192--203.

\bibitem{nguyen2019neural}
Nguyen, A.~T., Tran, N.~D., Vu, T.~T., Pham, T.~D., Vu, Q.~T., and Han, J.~H.,
  2019, \enquote{A Neural-network-based Approach to Study the Energy-optimal
  Hovering Wing Kinematics of a Bionic Hawkmoth Model,}
  \hrefurl{https://doi.org/10.1007/S42235-019-0105-5}{Journal of Bionic
  Engineering}, \textbf{16}, pp. 904--915.

\bibitem{bhat2020effects}
Bhat, S.~S., Zhao, J., Sheridan, J., Hourigan, K., and Thompson, M.~C., 2020,
  \enquote{Effects of flapping-motion profiles on insect-wing aerodynamics,}
  \hrefurl{https://doi.org/10.1017/JFM.2019.929}{Journal of Fluid Mechanics},
  \textbf{884}, p.~A8.

\bibitem{lang2023sensitivity}
Lang, X., Song, B., Yang, W., Yang, X., and Xue, D., 2023, \enquote{Sensitivity
  Analysis of Wing Geometric and Kinematic Parameters for the Aerodynamic
  Performance of Hovering Flapping Wing,}
  \hrefurl{https://doi.org/10.3390/AEROSPACE10010074}{Aerospace 2023, Vol. 10,
  Page 74}, \textbf{10}, p.~74.

\bibitem{terze2021optimized}
Terze, Z., Pandža, V., Kasalo, M., and Zlatar, D., 2021, \enquote{Optimized
  flapping wing dynamics via DMOC approach,}
  \hrefurl{https://doi.org/10.1007/S11071-020-06119-Y}{Nonlinear Dynamics},
  \textbf{103}, pp. 399--417.

\bibitem{sun2002lift}
Sun, M. and Tang, J., 2002, \enquote{{Lift and power requirements of hovering
  flight in Drosophila virilis},}
  \hrefurl{https://doi.org/10.1242/jeb.205.16.2413}{Journal of Experimental
  Biology}, \textbf{205}(16), pp. 2413--2427.

\bibitem{huang2019optimization}
Huang, M., 2019, \enquote{Optimization of flapping wing mechanism of bionic
  eagle,} \hrefurl{https://doi.org/10.1177/0954410018794339}{Proceedings of the
  Institution of Mechanical Engineers, Part G: Journal of Aerospace
  Engineering}, \textbf{233}(9), pp. 3260--3272.

\bibitem{han2023twist}
Han, Y.-J., Yang, H.-H., and Han, J.-H., 2023, \enquote{Twist-Coupled Flapping
  Mechanism for Bird-Type Flapping-Wing Air Vehicles,}
  \hrefurl{https://doi.org/10.1115/1.4062339}{Journal of Mechanisms and
  Robotics}, \textbf{15}(5), p. 051017.

\bibitem{nguyen2016extended}
Nguyen, A.~T., Kim, J.~K., Han, J.~S., and Han, J.~H., 2016, \enquote{Extended
  unsteady vortex-lattice method for insect flapping wings,}
  \hrefurl{https://doi.org/10.2514/1.C033456}{Journal of Aircraft},
  \textbf{53}, pp. 1709--1718.

\bibitem{yang2024parameter}
Yang, H.~H., Lee, S.~G., Addo-Akoto, R., and Han, J.~H., 2024,
  \enquote{Parameter Optimization of Foldable Flapping-Wing Mechanism for
  Maximum Lift,} \hrefurl{https://doi.org/10.1115/1.4056869/1156663}{Journal of
  Mechanisms and Robotics}, \textbf{16}.

\bibitem{rai2010unified}
Rai, A.~K., Saxena, A., and Mankame, N.~D., 2010, \enquote{Unified synthesis of
  compact planar path-generating linkages with rigid and deformable members,}
  \hrefurl{https://doi.org/10.1007/S00158-009-0458-1}{Structural and
  Multidisciplinary Optimization}, \textbf{41}, pp. 863--879.

\bibitem{dickinson1999wing}
Dickinson, M.~H., Lehmann, F.~O., and Sane, S.~P., 1999, \enquote{Wing rotation
  and the aerodynamic basis of insect right,}
  \hrefurl{https://doi.org/10.1126/SCIENCE.284.5422.1954}{Science},
  \textbf{284}, pp. 1954--1960.

\bibitem{persson2012numerical}
Persson, P.-O., Willis, D., and Peraire, J., 2012, \enquote{Numerical
  simulation of flapping wings using a panel method and a high-order
  Navier–Stokes solver,}
  \hrefurl{https://doi.org/10.1002/nme.3288}{International Journal for
  Numerical Methods in Engineering}, \textbf{89}(10), pp. 1296--1316.

\bibitem{ansari2006non1}
Ansari, S.~A., Żbikowski, R., and Knowles, K., 2006, \enquote{Non-linear
  unsteady aerodynamic model for insect-like flapping wings in the hover. Part
  1: Methodology and analysis,}
  \hrefurl{https://doi.org/10.1243/09544100JAERO49}{Proceedings of the
  Institution of Mechanical Engineers, Part G: Journal of Aerospace
  Engineering}, \textbf{220}(2), pp. 61--83.

\bibitem{ansari2006non2}
Ansari, S.~A., Żbikowski, R., and Knowles, K., 2006, \enquote{Non-linear
  unsteady aerodynamic model for insect-like flapping wings in the hover. Part
  2: Implementation and validation,}
  \hrefurl{https://doi.org/10.1243/09544100JAERO50}{Proceedings of the
  Institution of Mechanical Engineers, Part G: Journal of Aerospace
  Engineering}, \textbf{220}(3), pp. 169--186.

\bibitem{roccia2013modified}
Roccia, B.~A., Preidikman, S., Massa, J.~C., and Mook, D.~T., 2013,
  \enquote{Modified unsteady vortex-lattice method to study flapping wings in
  hover flight,} \hrefurl{https://doi.org/10.2514/1.J052262}{AIAA Journal},
  \textbf{51}, pp. 2628--2642.

\bibitem{wang2004unsteady}
Wang, Z.~J., Birch, J.~M., and Dickinson, M.~H., 2004, \enquote{{Unsteady
  forces and flows in low Reynolds number hovering flight:two-dimensional
  computations vs robotic wing experiments},}
  \hrefurl{https://doi.org/10.1242/jeb.00739}{Journal of Experimental Biology},
  \textbf{207}(3), pp. 449--460.

\bibitem{katz2001low}
Katz, J. and Plotkin, A., 2001, \textit{Low-speed aerodynamics}, Vol.~13,
  Cambridge university press.

\bibitem{polhamus1971predictions}
Polhamus, E.~C., 1971, \enquote{Predictions of vortex-lift characteristics by a
  leading-edge suctionanalogy,}
  \hrefurl{https://doi.org/10.2514/3.44254}{Journal of Aircraft},
  \textbf{8}(4), pp. 193--199.

\bibitem{bartlett1955experimental}
BARTLETT, G.~E. and VIDAL, R.~J., 1955, \enquote{Experimental Investigation of
  Influence of Edge Shape on the Aerodynamic Characteristics of Low Aspect
  Ratio Wings at Low Speeds,} \hrefurl{https://doi.org/10.2514/8.3391}{Journal
  of the Aeronautical Sciences}, \textbf{22}(8), pp. 517--533.

\bibitem{cao2016application}
Cao, Y., Yan, H., Liu, T., and Yang, J., 2016, \enquote{Application of
  Quasi-Monte Carlo Method Based on Good Point Set in Tolerance Analysis,}
  \hrefurl{https://doi.org/10.1115/1.4032909}{Journal of Computing and
  Information Science in Engineering}, \textbf{16}(2), p. 021008.

\bibitem{roshanbin2017colibri}
Roshanbin, A., Altartouri, H., Karásek, M., and Preumont, A., 2017,
  \enquote{COLIBRI: A hovering flapping twin-wing robot,}
  \hrefurl{https://doi.org/10.1177/1756829317695563}{International Journal of
  Micro Air Vehicles}, \textbf{9}(4), pp. 270--282.

\bibitem{lua2014scaling}
Lua, K.~B., Lim, T.~T., and Yeo, K.~S., 2014, \enquote{Scaling of Aerodynamic
  Forces of Three-Dimensional Flapping Wings,}
  \hrefurl{https://doi.org/10.2514/1.J052730}{AIAA Journal}, \textbf{52}(5),
  pp. 1095--1101.

\bibitem{ramasamy2007reynolds}
Ramasamy, M. and Leishman, J.~G., 2007, \enquote{A Reynolds Number Based Blade
  Tip Vortex Model,} \hrefurl{https://doi.org/10.4050/JAHS.52.214}{Journal of
  the American Helicopter Society}, \textbf{52}, pp. 214--223.

\bibitem{vatistas1991simpler}
Vatistas, G.~H., Kozel, V., and Mih, W., 1991, \enquote{A simpler model for
  concentrated vortices,}
  \hrefurl{https://doi.org/10.1007/BF00198434}{Experiments in Fluids},
  \textbf{11}(1), pp. 73--76.

\bibitem{milnethomson}
Milne-Thomson, L.~M., 1968, \textit{Theoretical Hydrodynamics}, 5th ed.,
  Macmillan, London.

\bibitem{lua2010aerodynamic}
Lua, K.~B., Lai, K.~C., Lim, T.~T., and Yeo, K.~S., 2010, \enquote{On the
  aerodynamic characteristics of hovering rigid and flexible hawkmoth-like
  wings,} \hrefurl{https://doi.org/10.1007/S00348-010-0873-5}{Experiments in
  Fluids}, \textbf{49}, pp. 1263--1291.

\end{thebibliography}
	
	\beginsupplement
	
	\twocolumn[
	\begin{center}
			\textbf{ \LARGE{ Supplement \\[1em]}} \normalsize 
			\vspace*{1em}
	\end{center}
	]
	
	\section{2D-UVLM Convergence study and validation}

\begin{figure}[!h]
	\centering
	\subfloat[]{\includegraphics[scale=1]{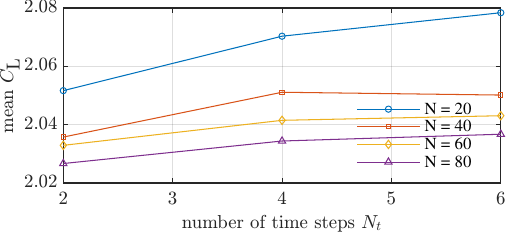}	}
	\vskip 1 em
	\subfloat[]{ \includegraphics[scale=1]{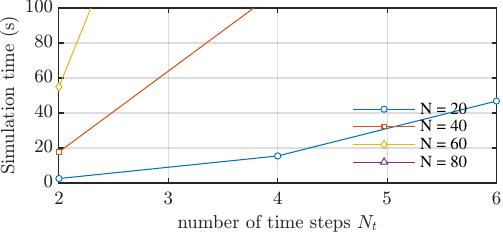} }
	\vskip 1em
	\subfloat[]{ \includegraphics[scale=1]{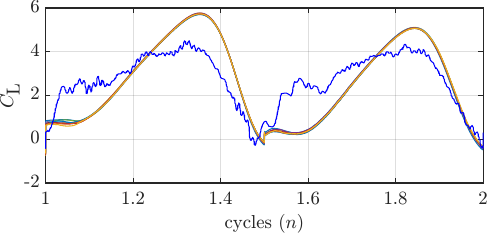} }
	\caption{ Convergence study of 2D UVLM: \textbf{(a)} Mean lift coefficients computed for harmonic kinematics data, for panels $N=20, 40, 60, 80$, and time-step divisor $N_\text{t} = 2,4,6$. \textbf{(b)} Computation time for each case compared. \textbf{(c)} Lift coefficients for $(N, N_\text{t}) = (20,2), (40,2), (40,4), (40,6), (60,2), (60,4), (60,6), (80,2), (80,4), (80,6) $ almost overlap over each other, indicating convergence. Experimental data (blue) \cite{lua2010aerodynamic}. Leading-edge suction force is not included here, as its convergence is implicitly satisfied. }
	\label{fig:uvlm_convergence}
\end{figure}

We varied the number of panels $N= 20, 40, 60, 80 $, and the nondimensional time step $\delta \tilde{t}  = \frac{\tilde{d}/ \tilde{\Omega}_\text{max}}{N_\text{t}}$ with $N_\text{t} = 2, 4 , 6 $, on harmonic kinematics data used by Nguyen \etal \cite{nguyen2016extended}, originally from experiments by Lua \etal \cite{lua2010aerodynamic} for their convergence study (Fig. \ref{fig:uvlm_convergence}). $N_\text{t} = 1$ yields the time for the lattice element to move by one element length $\tilde{d}$ with velocity $\tilde{\Omega}_\text{max}$. Among the 12 combinations of $N$ and $N_\text{t}$, 9 converged with the mean variation in lift coefficient $\epsilon = \delta \bar{C}_\text{L} = 0.025$, for $N=40, 60, 80$ and $N_\text{t} = 2, 4, 6 $. The computation time for two flapping cycles, however, varied significantly from 18 sec for $N=40$, $N_\text{t} = 2$, to 1175 sec for $N=60$, $N_\text{t} = 6$, on a Ryzen 5600X processor with 32 GB DDR4 RAM, running C++ MEX code in parallel with OpenMP in MATLAB 2024a. We also observed that for $N=20$, $N_\text{t} = 2$, the lift coefficient can be considered to practically converge within $\epsilon = 0.03$. This takes only 2.5 s, and is approximately 7 times faster than the next fastest in the set. For optimizations, we need fast computations. Therefore, we chose $N=20$ and $N_\text{t} = 2$ for all our simulations. We observed that for Squire's parameter $a=10^{-1}$, convergence was achieved in all cases considered in this study. Figure \ref{fig:uvlm_convergence} shows that lift coefficients for the converged and selected parameters almost overlap. 

\begin{figure*}[!h]
	\centerfloat
	
	\subfloat[Lift coefficient: 3D-UVLM ]{\includegraphics[scale=1]{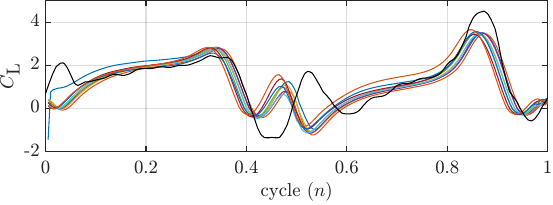}}  
	\hspace*{1em}
	\subfloat[Drag coefficient: 3D-UVLM ]{\includegraphics[scale=1]{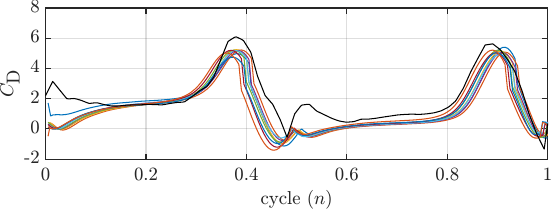} }
	\\
	\vspace*{1em}
	\\
	\subfloat[Lift coefficient: 2D-UVLM ]{\includegraphics[scale=1]{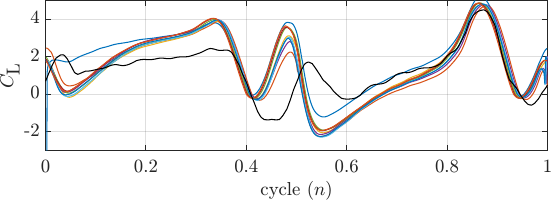} }
	\hspace*{1em}
	\subfloat[Drag coefficient: 2D-UVLM ]{\includegraphics[scale=1]{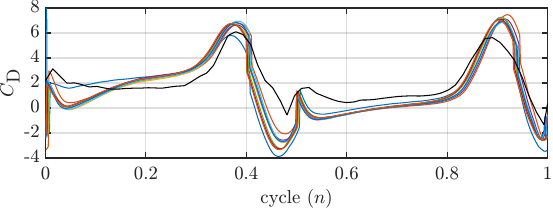}}
	\\
	\vspace*{1em}
	\\
	\subfloat[Mean Lift coefficient ]{\includegraphics[scale=1]{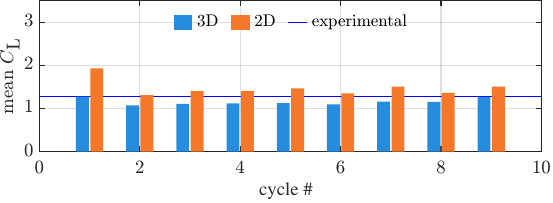}} 
	\hspace*{1em}
	\subfloat[Mean Drag coefficient ]{\includegraphics[scale=1]{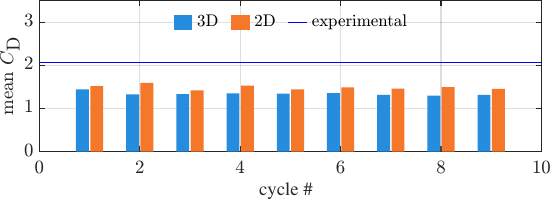}}

	\caption{Comparison of 2D and 3D-UVLM: (a)--(d) Lift and drag coefficients for the fruit fly wing kinematics for nine cycles, overlapped to assess the periodicity. Curve in black - experimental data \cite{dickinson1999wing}. (e)--(f) Mean values over each cycle compared. Lift estimates are close, while drag is underestimated compared to the experimental data. Due to the slow diffusion of wake vortices in 2D-UVLM, the periodicity was affected. To circumvent that, we truncated the number of wake vortices to 1.5 flapping cycles. Wake vortices older than that were merged in a blob at infinity ( $(10^8,0)$ ), to maintain Kelvin circulation conservation. From (e)--(f), we can say that, for 2D-UVLM, the lift and drag coefficients are nearly periodic from the second cycle. For 3D-UVLM, they are periodic from the second cycle, too.}
	\label{fig:UVLM_2d_3d_periodicity}
\end{figure*}

We observed a nonperiodic trend in the lift coefficient during 2D-UVLM simulations for many flapping cycles (>2). The mean lift coefficient drops over consecutive cycles. This happens because the wake cannot diffuse in the spanwise direction in 2D, unlike in 3D, which results in low bound vortex strengths. Therefore, we limited wake vortices to a maximum number. When that number is reached, we merge older wake vortices into a blob at infinity - numerically at $(10^8, 0)$. Their time step is set to that of the first wake vortex, which is immaterial because of the large distance. Effectively, they produce no effect -- a similar effect of faster diffusion like in real flows. This procedure results in periodic lift coefficients (Fig. \ref{fig:UVLM_2d_3d_periodicity}) from the second cycle onwards. It also significantly reduces the computation time for many flapping cycles. We tested the maximum number with different values. If the number is too large, we have the same problem of reduced lift coefficients. If it is too small, it means we diffuse them too fast. It results in unrealistically high mean lift coefficients and reduced drag coefficients. With trial and error, we found that number corresponded to time-steps in 1.5 flapping cycles. 

\section{Comparison of 2D and 3D UVLM}
\label{sec:app:2d_3d}

We also implemented 3D UVLM with leading-edge suction computed similarly using the Kutta Joukowski expression for leading-edge panels. The coefficient $\eta$ for 2D and 3D was set to 0.2. Convergence analysis resulted in the use of $ = M=9$ chordwise and $N=15$ spanwise panels on nondimensionalized wing planform, with 100 time-steps per flapping cycle, with a mean lift coefficient in the range of 0.03, compared with $M=24, N=48$ and 400 time-steps. The computation took approximately 1.8 seconds per cycle for harmonic flight kinematics. However, we need a finer mesh to get a better spatial distribution of aerodynamic load. $M=12, N=18$ with 100 time-steps consumed 14 s for two cycles. We need two cycles, as the lift and drag coefficients obtained are periodic from the second cycle onwards (Fig. \ref{fig:UVLM_2d_3d_periodicity}). Since we compute only lift and drag coefficients in this supplementary material, a coarser mesh with $M=9, N=15$ with 100 time-steps is sufficient (from the convergence study: not shown here), and is used for all results herein.

\begin{figure*}[!h]
	\centerfloat
	
	\subfloat[Lift coefficient: Oval-A]{\includegraphics[scale=1]{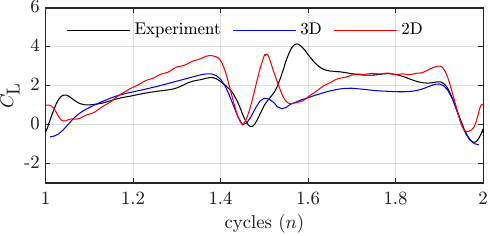}}  
	\hspace*{1em}
	\subfloat[Drag coefficient: Oval-A]{\includegraphics[scale=1]{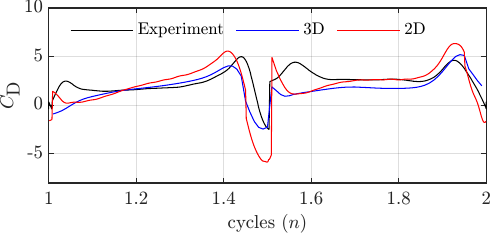} }
	\\
	\vspace*{1em}
	\\
	\subfloat[Lift coefficient: Oval-B]{\includegraphics[scale=1]{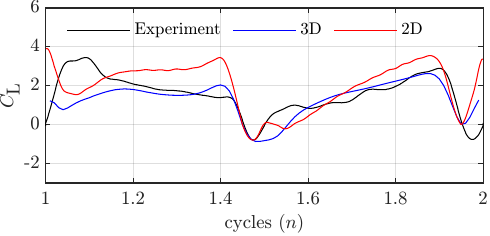} }
	\hspace*{1em}
	\subfloat[Drag coefficient: Oval-B]{\includegraphics[scale=1]{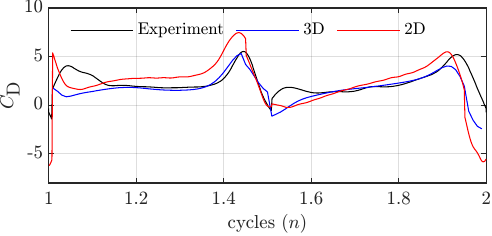}}
	\\
	\vspace*{1em}
	\\
	\subfloat[Lift coefficient: Figure of 8-A]{\includegraphics[scale=1]{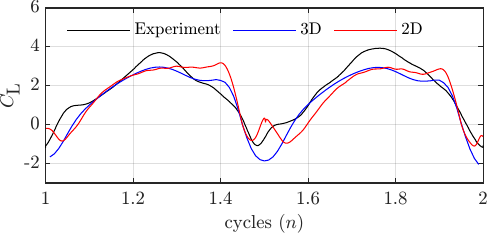}} 
	\hspace*{1em}
	\subfloat[Drag coefficient: Figure of 8-A]{\includegraphics[scale=1]{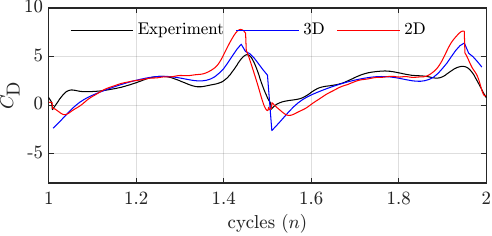}}
	\\
	\vspace*{1em}
	\\
	\subfloat[Lift coefficient: Figure of 8-B]{\includegraphics[scale=1]{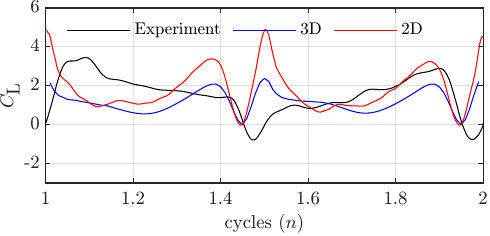} }
	\hspace*{1em}
	\subfloat[Drag coefficient: Figure of 8-B]{\includegraphics[scale=1]{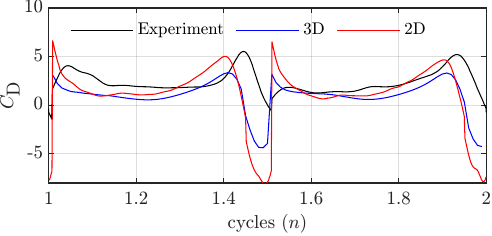}}
	\\
	\vspace*{1em}
	\\
	\subfloat[Lift coefficient: Fruit fly]{\includegraphics[scale=1]{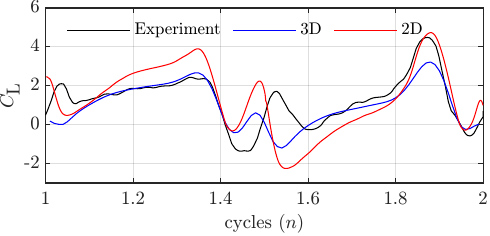}} 
	\hspace*{1em}
	\subfloat[Drag coefficient: Fruit fly]{\includegraphics[scale=1]{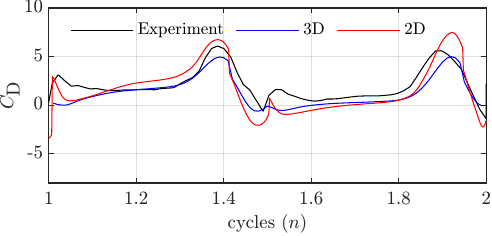}}
	\\
	
	\caption{Lift and drag coefficients from 2D-UVLM (thick and blue) for oval, figure-8, and fruit fly flight patterns. Experimental data (thin and black) from oval and figure-8 patterns are taken from \cite{sane2001control} and for fruit fly pattern from \cite{dickinson1999wing}.}
	\label{fig:fig8_Oval}
\end{figure*}

\begin{figure}[!h]
	\centering
	\subfloat[]{\includegraphics[scale=1]{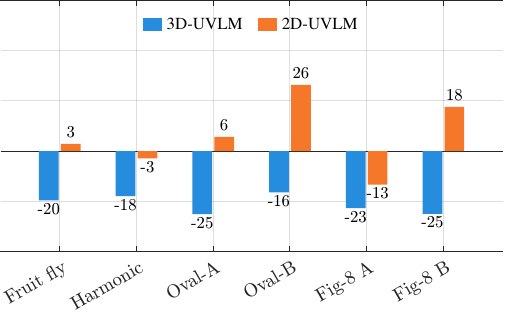}	}
	\vskip 1 em
	\subfloat[]{ \includegraphics[scale=1]{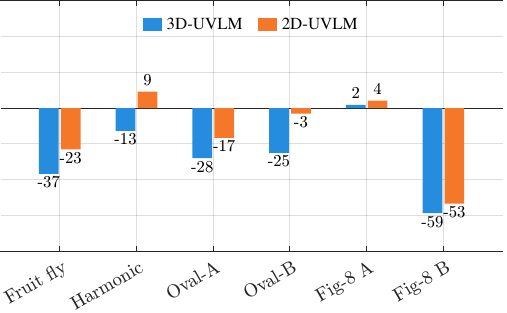} }
	\caption{ Percentage deviation of mean lift coefficients in \textbf{(a)} and mean drag coefficients in \textbf{(b)} from experimental data for flapping patterns - fruit fly \cite{dickinson1999wing}, harmonic \cite{lua2010aerodynamic}, Oval-A, Oval-B, figure-8 A, and figure-8 B\cite{sane2001control}, computed using 2D and 3D UVLM. The coefficient $\eta$ for leading edge suction is taken as 0.2.  }
	\label{fig:uvlm_2D_3D_compare}
\end{figure}

For further comparisons, the fruit fly wing planform was used for fruit fly, Oval and figure-8 flight patterns \cite{dickinson1999wing, sane2001control}, and hawkmoth wing planform for harmonic flight kinematics \cite{lua2010aerodynamic}. Figures \ref{fig:uvlm_2D_3D_compare} and \ref{fig:fig8_Oval}  compare the lift and drag coefficients computed from 2D-UVLM and 3D-UVLM with experimental data. We observed that both versions captured the lift and drag trends well. Some peaks and valleys in 2D-UVLM were overestimated, and some were underestimated in 3D-UVLM. The mean lift and drag coefficients show similar deviations from the experimental values, indicating no clear advantage of 3D-UVLM. However, the 2D-UVLM performed about 5 times faster than the 3D-UVLM - providing a significant advantage in optimizations where many function evaluations are required. It also helps significantly in the robustness analysis for the same reason.

\section{Variation of lift and drag coefficients with flapping frequency}

By varying the flapping frequency between 10--20 Hz, we computed mean lift and drag coefficients for different cases listed in Tab. \ref{tab:lift_drag_frequency}. The fruit fly wing pitching angle profile was used for all cases \cite{dickinson1999wing}. The coefficients do not vary much, so we can assume them to be constant in that frequency range. 
\begin{table}[!h]
	\centering
	\caption{Mean lift and drag coefficients for varied flapping frequencies. }
	\label{tab:lift_drag_frequency}
	\begin{tabular}{l c c c c} 
		\hline 
		\hline 
		\addlinespace[3pt] 
		& 	 & 10 Hz 				& 	 15 Hz 	 & 	 20 Hz  \\ 
		\addlinespace[3pt] 
		\hline 
		\addlinespace[3pt] 
		Deng \etal \cite{deng2021design} 	 & $\bar{C}_\text{L}$ 	 & 1.558 	 & 1.560 	 & 1.555 \\ 
		& $\bar{C}_\text{D}$ 	 & 2.341 	 & 2.335 	 & 2.332 \\ 
		\addlinespace[3pt] 
		Karásek \etal \cite{karásek2014pitch} 	 & $\bar{C}_\text{L}$ 	 & 1.431 	 & 1.429 	 & 1.430 \\ 
		& $\bar{C}_\text{D}$ 	 & 2.200 	 & 2.202 	 & 2.203 \\ 
		\addlinespace[3pt] 
		Coleman \etal \cite{coleman2017development}  & $\bar{C}_\text{L}$ 	 & 1.758 	 & 1.747 	 & 1.754 \\ 
		& $\bar{C}_\text{D}$ 	 & 2.355 	 & 2.350 	 & 2.353 \\ 
		\addlinespace[3pt] 
		\hline 
		\hline 
	\end{tabular} 
\end{table}

\section{Selected optimal mechanisms compared with existing ones}
\begin{figure}[!h]
	\centering
	\includegraphics[scale=1]{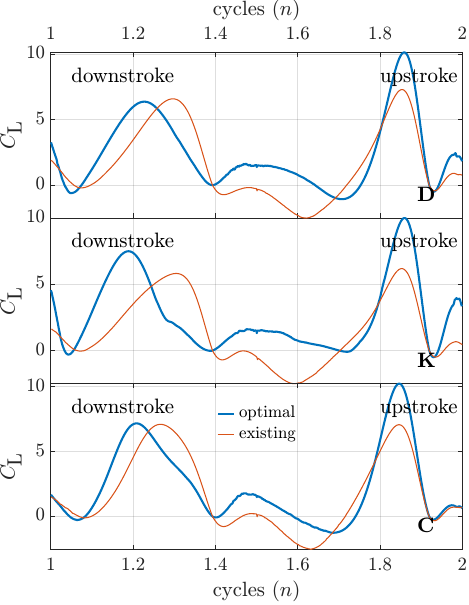}					
	\caption{ Lift coefficients $C_\text{L}$ generated by the existing mechanisms in thin red; selected Pareto optimal mechanisms in thick blue: topologies used: D - Deng \etal \cite{deng2021design}, K - Karásek \etal \cite{karásek2014pitch}, C - Coleman \etal \cite{coleman2017development}. The optimal mechanisms have smaller negative valleys during upstrokes. Peaks due to rotational effect (around $n = 1.9$) during upstrokes are higher when sweeping velocities are higher than those of the existing mechanisms (Fig. \ref{fig:optimalKinematics}). Peaks due to wing--wake interaction (around $n=1.5$) at the end of downstroke are higher. The optimal mechanisms sweep much more slowly around these times. During downstrokes, the lift develops faster, coinciding with the faster acceleration of the optimal mechanisms in these periods (around $n=1.2$).}
	\label{fig:compareLifts}
\end{figure}

Figure \ref{fig:compareLifts} compares the lift coefficients of selected optimal mechanisms with existing ones.

\end{document}